\documentclass[11pt]{article}

\usepackage[final]{acl}

\usepackage{times}
\usepackage{latexsym}

\usepackage[T1]{fontenc}
\usepackage[utf8]{inputenc}

\usepackage{microtype}

\usepackage{inconsolata}

\usepackage{graphicx}

\usepackage{booktabs}          
\usepackage{xcolor}                       % \textcolor for \cmark/\xmark
\usepackage{tcolorbox}                    % promptbox environment
\usepackage{tikz}                         % causal-graph diagrams
\usetikzlibrary{arrows.meta,positioning,shapes,decorations.pathreplacing}
\usepackage{pifont}                       % \ding for \cmark/\xmark
\usepackage{amsmath}
\usepackage{fontawesome5}

\usepackage{float}

\newcommand{\projectname}{ReCITE}
\newcommand{\acryonymdefinition}{Real-world CausalIty from Textual Evidence}

\newtcolorbox{promptbox}[1]{colback=gray!5!white,colframe=gray!75!black,
   fonttitle=\bfseries\scriptsize,fontupper=\ttfamily\footnotesize,
   title=#1}

\tcbuselibrary{skins,breakable}
\newtcolorbox{longpromptbox}[1]{enhanced,breakable,
  colback=gray!5!white,colframe=gray!75!black,
  fonttitle=\bfseries\scriptsize,fontupper=\ttfamily\footnotesize,
  title=#1,
  before skip=10pt,after skip=10pt,
  sharp corners=south}

\definecolor{pricecolor}{RGB}{204, 102, 0}     % Orange
\definecolor{buyercolor}{RGB}{51, 102, 153}    % Blue
\definecolor{revenuecolor}{RGB}{34, 139, 34}   % Green

\newcommand{\cmark}{\textcolor{green!60!black}{\ding{51}}}
\newcommand{\partialmarktext}{\textcolor{yellow!80!black}{\faExclamationTriangle}}
\newcommand{\xmark}{\textcolor{red!80!black}{\ding{55}}}%

\title{Can Large Language Models Infer Causal Relationships\\ from Real-World Text?}

\author{
  Ryan Saklad$^{1}$ \quad
  Aman Chadha$^{2}$\thanks{Work done outside position at Apple.} \quad
  Oleg Pavlov$^{1}$ \quad
  Raha Moraffah$^{1}$ \\
  $^{1}$Worcester Polytechnic Institute \quad $^{2}$Apple
}

\begin{document}
\maketitle
\begin{abstract}
Understanding and inferring causal relationships from texts is a core aspect of human cognition and is essential for advancing large language models (LLMs) towards artificial general intelligence. Existing work evaluating LLM causal reasoning primarily relies on synthetic or simplified texts with explicitly stated causal relationships. These texts typically feature short passages and few causal relations, failing to reflect the complexities of real-world reasoning. In this paper, we investigate whether LLMs are capable of inferring causal relationships from real-world texts. We develop a benchmark drawn from real-world academic literature, which includes diverse texts with respect to length, complexity (different levels of explicitness, number of causal events and relationships), and domain. To the best of our knowledge, our benchmark is the first-ever real-world dataset for this task. Our experiments on this dataset show that LLMs face significant challenges in inferring causal relationships from real-world text, with the best-performing model achieving an average F$_1$ score of only 0.535. Through systematic analysis across aspects of real-world text (explicitness, number of causal events and relationships, length of text, domain), our benchmark offers targeted insights for further research into advancing LLM causal reasoning. Our code and dataset can be found at \url{https://github.com/Ryan-Saklad/ReCITE}.
\end{abstract}

\section{Introduction}

\begin{figure}[!t]
\captionsetup{skip=5pt}  % tighten space below the caption
\centering
\resizebox{\columnwidth}{!}{%
\begin{tikzpicture}[
    >={Stealth[round]},
    % ---------- styles ----------
    var/.style = {circle, draw=black,
                  minimum size=2.2cm,            % smaller bubbles
                  font=\large\bfseries,          % node text
                  align=center, inner sep=3pt},
    textbox/.style = {rectangle, draw=white!80, rounded corners,
                      font=\normalsize, fill=gray!5,
                      text width=5.1cm,          % identical widths
                      inner sep=6pt, anchor=north},
    edge/.style = {->, thick}
]

% Side-by-side explanatory text
\node[textbox] (implicit) at (-3,4.6) {%
  \parbox[c][3.8cm][c]{5.1cm}{\centering
  \underline{\textbf{Implicit}}\\[0.5em]
  
  Athletes spent more time \textcolor{pricecolor}{\textbf{training}} at the 
  training facility this season. \ldots\ Coaches noticed improved 
  \textcolor{buyercolor}{\textbf{attitudes}} during warmups. 
  \ldots\ Competition \textcolor{revenuecolor}{\textbf{results}} exceeded 
  expectations across the board.
  }};

\node[textbox] (explicit) at (3,4.6) {%
  \parbox[c][3.8cm][c]{5.1cm}{\centering
  \underline{\textbf{Explicit}}\\[0.5em]
  
  More \textcolor{pricecolor}{\textbf{practice}} directly improves 
  \textcolor{revenuecolor}{\textbf{performance}} through skill development. 
  \textcolor{pricecolor}{\textbf{Practice}} also builds 
  \textcolor{buyercolor}{\textbf{confidence}}. Greater 
  \textcolor{buyercolor}{\textbf{confidence}} leads to better 
  \textcolor{revenuecolor}{\textbf{performance}}.}};

\node[var, fill=pricecolor!20, draw=pricecolor!80] (practice) at (0,-0.9) {Practice};
\node[var, fill=buyercolor!20, draw=buyercolor!80] (confidence) at (-4.5,-2.2) {Confidence};
\node[var, fill=revenuecolor!20, draw=revenuecolor!80] (performance) at (4.5,-2.2) {Performance};

\draw[edge] (practice)   -- (confidence);
\draw[edge] (practice)   -- (performance);
\draw[edge] (confidence) -- (performance);

\end{tikzpicture}}
\caption{An example causal graph illustrating the difference between explicit and implicit texts describing the same causal relationships. The explicit text directly states causal relationships using clear language (``directly improves,'' ``leads to''). The implicit text describes the same relationships without stating causation (``spent more time training,'' ``noticed improved attitudes''). This exemplifies a key challenge with real-world texts, where causal reasoning must be used to construct the graph.}
\vspace{-16pt}
\label{fig:example-cg}
\end{figure}

The ability to identify and understand causal relationships embedded within texts is a fundamental aspect of human intelligence \cite{pearl2009causality, gopnik2004causal} and is crucial for complex decision-making. Humans excel at inferring these relationships from text, even when they are not explicitly stated \cite{graesser1994inferences}. As a simplified example, given the text ``Athletes spent extra time at the training facility this season. \ldots\ Coaches noticed improved attitudes during warmups. \ldots\ Competition results exceeded expectations across the board,''\footnote{Ellipses indicate omitted text between sentences, representing how real-world documents intersperse causal information throughout lengthy passages.} humans identify the underlying causal events and infer the relationships: practice causes improved performance, practice causes increased confidence, and confidence causes better performance. Notably, none of these causal relationships need to be directly stated for a human to infer them.

Given the importance of causality to intelligence, significant research has focused on assessing the causal reasoning capabilities of large language models (LLMs). Much of this work evaluates LLMs' stored knowledge about causal concepts \cite{zhou2024causalbenchcomprehensivebenchmarkcausal, kıcıman2024causalreasoninglargelanguage, miliani2025explicaevaluatingexplicitcausal, joshi2024coldcausalreasoningclosed}, rather than their ability to infer causal relationships from text. While some studies do examine extracting causal relationships from text, they often use synthetically generated or simplified texts with explicitly stated causal relations~\cite{Veldhuis2024, hosseinichimeh2024, Oh2025, Jin2024, joshi2024coldcausalreasoningclosed, lasheras-pinheiro-2025-calquest}. This falls short of real-world scenarios where causal relationships are embedded in lengthy, complex texts with varying degrees of explicitness and diverse domains. Figure~\ref{fig:example-cg} illustrates this contrast: while synthetic texts may present clear, explicit causal statements, real-world texts convey causal relationships implicitly, requiring genuine causal reasoning to extract the underlying structure.

To address this gap, we introduce \acryonymdefinition~(\textbf{\projectname}), a novel benchmark drawn from real-world academic literature. {\projectname} features diverse texts varying in length, explicitness, complexity (5--140 causal events, 6--205 relationships), and spanning multiple domains. To enable automated evaluation at scale, we develop an LLM-as-a-Judge framework for assessing generated causal graphs. Experiments with state-of-the-art (SOTA) LLMs on {\projectname} demonstrate that LLMs struggle to infer causal relationships from real-world text, with the best models achieving an average F$_1$ score of only 0.535. Our results show that LLMs particularly struggle with implicitly stated causal information, with performance degrading substantially as explicitness decreases. By  characterizing these deficiencies across multiple dimensions, our benchmark offers targeted insights for advancing LLM causal reasoning. Our main contributions are as follows:
\begin{itemize}
    \item We propose {\projectname}, the first benchmark to evaluate real-world causal reasoning abilities of LLMs from text, featuring samples with realistic variation in length, explicitness, causal complexity, and domain.
    \item We conduct extensive experiments on ten SOTA LLMs, revealing significant limitations in LLMs' ability to perform causal reasoning from real-world text, with the best-performing model achieving an F$_1$ score of only 0.535. We also show that performance is poor even when identification of causal events is removed, indicating causal reasoning is the primary challenge.
    \item We develop a novel LLM-as-a-Judge evaluation framework that enables automated assessment of generated graphs.
\end{itemize}

\section{Related Work}
\label{sec:related-work}

\paragraph{Causal Reasoning with LLMs}

As LLMs have grown in capabilities, significant research has explored their causal reasoning abilities under varied evaluation styles (see surveys by Kıcıman et al.~\cite{kıcıman2024causalreasoninglargelanguage} and Wan et al.~\cite{wan2025largelanguagemodelscausal}). Some benchmarks focus on predicting causal structure from a provided variable set, largely bypassing long-form textual inference (e.g., CausalBench \cite{zhou2024causalbenchcomprehensivebenchmarkcausal}), while others utilize textual inputs but rely on synthetic or simplified narratives with explicit, often pairwise, relationships or small graphs \cite{chen2024causalevaluationlanguagemodels, Jin2024, miliani2025explicaevaluatingexplicitcausal, hosseinichimeh2024}.
A complementary line of work targets sentence-level causal language in real-world corpora \cite{ding-etal-2025-multi} or evaluates causal understanding via question-answering over short contexts \cite{lasheras-pinheiro-2025-calquest, chi2025unveilingcausalreasoninglarge, joshi2024coldcausalreasoningclosed}.
While these benchmarks are valuable, they fail to reflect real-world causal reasoning involving lengthy texts, large causal graphs, and varied levels of explicitness. Unlike prior benchmarks that rely on synthetic texts, short contexts, pairwise relations, or provided variable lists (see Appendix~\ref{sec:detailed-comparison}), {\projectname} requires full graph construction from lengthy real-world documents where causal relationships must be inferred rather than extracted from explicit statements.

\paragraph{Knowledge Graph Discovery with LLMs}
Recent work has applied LLMs to knowledge graph construction \cite{Bratanic2024, Zhang2024, Yu2023}, and related analyses in economics use language models to map concept-relationship graphs across large paper corpora \cite{garg2025causalclaimseconomics}. However, these approaches emphasize factual relations or coarse claim links, differing from the challenge in {\projectname} of reconstructing the full causal graph from long-form text.

\section{The {\projectname} Benchmark}
\label{sec:dataset-creation}

We present \acryonymdefinition~(\textbf{\projectname}), to the best of our knowledge, the first benchmark for realistic causal discovery from text. Given an input text, the task is to construct a causal graph where nodes represent causal events and directed edges represent their causal relationships. Constructing a realistic benchmark for this task requires source texts that exhibit real-world complexity and reliable ground-truth causal graphs. Academic papers employing causal loop diagrams satisfy both requirements: they contain detailed textual descriptions of goals, methodology, background, and causal reasoning without relying on non-textual elements, and each paper centers around a single, human-authored causal graph that articulates its modeling assumptions. This enables accurate annotation of ground-truth graphs, and makes construction from the text feasible even when elements are not mentioned explicitly. We construct {\projectname} through a systematic 3-step pipeline: (i) Academic paper collection, (ii) Causal Graph Annotation, (iii) Post-processing.

\subsection{Academic Paper Collection}
\label{sec:paper-identification}

We utilize API access to academic paper repositories, restricting our search to open access sources (MDPI and PLOS) to ensure legal redistribution of texts and diagrams. We use a keyword search for the term ``causal loop diagram,'' resulting in 646 candidate papers. We then manually exclude all papers with graphs unsuitable for annotation, including ones with no causal graph or multiple unrelated causal graphs, purely illustrative or non-causal diagrams, and causal graphs with poor legibility or ambiguous graph elements. We retain only papers containing a single causal graph (either as the sole graph, or as the culmination of a sequence of graphs) so that it can serve as ground-truth. During manual filtering, the location of each primary causal graph is labeled to ensure accurate annotation (e.g., ``Top diagram on page 7'').

Table~\ref{tab:dataset-statistics} displays summary statistics of the resulting corpus across key dimensions. Text lengths range from 684 to 171,141 characters. Explicitness varies substantially, with 87.7\% of causal events mentioned in the text. The corpus also spans 17 distinct fields: external classification via OpenAlex Topics~\cite{priem2022openalexfullyopenindexscholarly} reveals that only 5.5\% of papers fall under core Economics, Econometrics \& Finance, with Environmental Science (20.2\%), Engineering (20.2\%), and Business/Management (12.0\%) comprising the largest shares. Embedding analysis confirms greater similarity to STEM literature than traditional economics (MMD $= 0.114$ vs.\ $0.141$; Appendix~\ref{app:textual-diversity}). Semantic diversity is similarly high: across 6,378 unique causal concepts, 92.9\% appear in only a single paper, and 97.6\% of paper pairs share zero causal concepts. Full graph topology statistics, including density, cyclicity, and motif distributions, are reported in Appendix~\ref{app:graph-diversity}.

\begin{table}[t]
   \centering
   \resizebox{\columnwidth}{!}{%
   \begin{tabular}{lcc}
   \toprule
       \textbf{Attribute} & \textbf{Mean ± SD} & \textbf{Range} \\
   \midrule
       Number of Samples & -- & 292 \\
       Text Length & 40541 ± 17722 & (684, 171141) \\
       Nodes per Graph & 25.0 ± 15.8 & (5, 140) \\
       Edges per Graph & 37.4 ± 24.3 & (6, 205) \\
       Explicitness per Node & 0.877 & -- \\
   \bottomrule
   \end{tabular}}
   \caption{{\projectname} dataset statistics showing its real-world attributes across key dimensions. Text length is measured in characters. Explicitness per node reports the proportion of nodes mentioned in the sample's text.}
   \vspace{-12pt}
   \label{tab:dataset-statistics}
\end{table}

\subsection{Causal Graph Annotation}
\label{sec:causal-graph-annotation}
As {\projectname} is a text-based benchmark, it is important that ground-truth causal graphs are converted from images to a text-based representation to serve as the ground-truth answer. We find through manual evaluation that vision-based LLMs are unsuitable for annotation due to frequent hallucinations, consistent with prior work \cite{white2025vlm_systemmaps, bai2025hallucinationmultimodallargelanguage}. Therefore, we employ annotators to convert each causal graph into a standardized format (``source\_variable'' -> ``sink\_variable''). They are provided detailed instructions for the annotation process, including which graph elements to include, a standardized output format, and step-by-step examples.

\subsection{Post-Processing}
\label{sec:post-processing}

\paragraph{Graph Post-Processing}
\label{sec:graph-processing}

To ensure the reliability of ground-truth graphs and mitigate potential human error, we utilize a rigorous post-processing pipeline. We first utilize code-based approaches to identify formatting mistakes and string matching approaches to automatically correct them. Any samples that are unable to be automatically corrected are flagged and manually corrected.

To verify transcription quality, a second annotator relabeled 37 randomly chosen diagrams ($879$ edges, $674$ nodes; $\approx13$\% of {\projectname}). We observed $22$ missing and $5$ spurious edges, giving edge–level precision~$=0.994$, recall~$=0.975$, $F_{1}=0.984$, and Cohen’s~$\kappa=0.987$.  Node labels showed 8 auto-correctable typos (e.g., \texttt{carbondioxide}) and 4 minor prefix/suffix omissions; no major name mismatches occurred. Full per-graph statistics are provided in Appendix~\ref{app:inter-annotator}. We explore utilization of code-based approaches for correction of node naming, but find that it is prone to erroneously combining distinct nodes (e.g., GDP and GNP). To address this, we utilize LLMs for automated correction (see Appendix~\ref{app:variable-correction}).

\paragraph{Text Post-Processing}
\label{sec:text-processing}

To convert PDF papers into a textual format suitable for LLM input, which cannot be done trivially with automated parsing \cite{Meuschke_2023}, we utilize a multi-step pipeline. First, we use the Python library PyMuPDF to extract the raw text for each paper. The output of this step contains numerous formatting errors, such as arbitrary line breaks. We utilize a multi-step LLM pipeline, due to the infeasibility of code-based approaches. Manual testing shows current LLMs cannot perform this process accurately in one step. We first prompt Mistral Small \cite{mistral2025} to convert from the PDF text to well-structured markdown. The goal of this step is to remove non-textual elements (which are impossible to accurately represent in markdown) and entirely irrelevant elements (to streamline the task and save on computational costs). Therefore, Mistral is tasked with outputting the markdown auto-regressively while skipping non-textual elements (such as images, charts, or other figures), in-line citations, references, publication information, and appendices.

{\color{blue}
\begin{tcolorbox}[colback=blue!10, colframe=blue, title=Normalization Tool]
\begin{verbatim}
{ 
 "normalizations": [
   {"start": "text to find (start)",
    "end": "text to find (end)",
    "replacement": "new text"},
    ... ]
}
\end{verbatim}
\end{tcolorbox}
}

However, this may include information that makes the task trivial (such as a table including all of the sample's causal relations), or it may erroneously reference removed elements (resulting in an internally inconsistent document). To correct these issues, we utilize o3-mini \cite{OpenAI2025} to remove explicit references to the causal graph and correct any references to missing elements using a normalization tool. Explicit references to the graph are unrealistic (such as a table listing every connection in the graph), as the task is to create a causal graph that does not already exist. However, we are careful to not remove other information about the graph to avoid making it unidentifiable. To do this, we provide the model with a normalization tool to minimize output tokens and unnecessary changes, while allowing for code-based validation.

\section{Experiments}
\label{sec:experiments}

In this section, we examine the performance of SOTA LLMs on {\projectname}. Specifically, we investigate the following research questions: (i) How well do current LLMs perform on real-world causal reasoning from text? (ii) How does performance vary across different characteristics of {\projectname}, including text length, explicitness, number of nodes and edges, and domain? Finally, we demonstrate common reasoning failures via a case study.

\subsection{Experiment Setup}
\label{sec:experiment-setup}

\paragraph{Models.}
\label{sec:models}

We evaluate diverse SOTA models spanning reasoning and instruction-tuned categories, proprietary and open-weight, as summarized in Table~\ref{tab:models}. We assess in the zero-shot setting and use default hyperparameters. We exclude samples where the model's maximum context length is exceeded (only Qwen 2.5 32B is affected, with 20 samples excluded). This applies across all experiments except where otherwise specified. For proprietary models, we retry responses where we do not receive an answer. We explicitly provide the expected number of nodes because the task is otherwise ill-defined; there are many valid levels of abstraction for each graph, and real-world tasks typically specify the desired level of granularity. We enforce strict JSON formatting for graph outputs to enable automated evaluation. For responses that fail formatting requirements, we apply a post-processing step using Mistral Small \cite{mistral2025} or GPT-5-mini \cite{openai2025gpt5} to convert malformed outputs into valid JSON while preserving the intended relationships. We use DeepSeek v3.2 \cite{deepseek2025v32} as our judge due to its long context, tractable cost, and status as a non-evaluated model.

\begin{table}[t]
  \centering
  \small
  \setlength{\tabcolsep}{2pt}
  \resizebox{\columnwidth}{!}{%
  \begin{tabular}{lcc}
    \toprule
    \textbf{Model} & \textbf{Type} & \textbf{Parameters} \\
    \midrule
    Claude Opus 4.5 \cite{anthropic2025claude45}       & Reasoning & N/A \\
    GPT 5.2 \cite{openai2025gpt52}                     & Reasoning & N/A \\
    Gemini 3 Pro \cite{google2025gemini3}              & Reasoning & N/A \\
    Gemini 3 Flash \cite{google2025gemini3}            & Reasoning & N/A \\
    GLM 4.7$^\dagger$ \cite{zhipu2025glm47}            & Reasoning & 355 \\
    Kimi K2$^\dagger$ \cite{moonshot2025k2}            & Reasoning & 1000 \\
    DeepSeek R1$^\dagger$ \cite{R12025}                & Reasoning & 671 \\
    QwQ 32B$^\dagger$ \cite{qwen2025qwq}               & Reasoning & 32 \\
    Qwen 2.5 32B$^\dagger$ \cite{qwen2024}             & Instruct & 32 \\
    Llama 3.1 8B$^\dagger$ \cite{grattafiori2024llama3herdmodels} & Instruct & 8 \\
    \bottomrule
  \end{tabular}%
  }
  \caption{Evaluated models. $^\dagger$Open-weight. Type denotes reasoning-tuned versus instruction-tuned. Parameters listed in billions, and N/A when not disclosed.}
  \vspace{-12pt}
  \label{tab:models}
\end{table}

\paragraph{Evaluation Metrics.}
\label{sec:evaluation-metrics}

\begin{table*}[ht]
   \centering
   \resizebox{\textwidth}{!}{%
   \begin{tabular}{lccccccc}
       \toprule
       \textbf{Model} & \textbf{Node Precision~($\uparrow$)} & \textbf{Node Recall~($\uparrow$)} & \textbf{Edge Precision~($\uparrow$)} & \textbf{Edge Recall~($\uparrow$)} & \textbf{F$_1$~($\uparrow$)} & \textbf{SHD~($\downarrow$)} & \textbf{Normalized SHD~($\downarrow$)} \\
       \midrule
       Claude Opus 4.5  & \(\mathbf{0.972 \pm 0.055}\) & \(\mathbf{0.571 \pm 0.215}\) & \(0.853 \pm 0.154\) & \(0.305 \pm 0.208\) & \(\mathbf{0.535 \pm 0.202}\) & \(\mathbf{47.0 \pm 36.5}\) & \(0.108 \pm 0.091\) \\
       GPT 5.2          & \(0.960 \pm 0.070\) & \(0.524 \pm 0.214\) & \(0.804 \pm 0.160\) & \(\mathbf{0.309 \pm 0.205}\) & \(0.510 \pm 0.201\) & \(58.5 \pm 45.1\) & \(0.130 \pm 0.094\) \\
       Gemini 3 Pro     & \(0.980 \pm 0.050\) & \(0.527 \pm 0.232\) & \(\mathbf{0.865 \pm 0.152}\) & \(0.281 \pm 0.215\) & \(0.501 \pm 0.221\) & \(48.3 \pm 37.2\) & \(0.114 \pm 0.101\) \\
       GLM 4.7          & \(0.970 \pm 0.085\) & \(0.524 \pm 0.226\) & \(0.808 \pm 0.169\) & \(0.274 \pm 0.199\) & \(0.493 \pm 0.210\) & \(50.8 \pm 37.8\) & \(0.116 \pm 0.089\) \\
       Kimi K2          & \(0.950 \pm 0.079\) & \(0.508 \pm 0.221\) & \(0.778 \pm 0.177\) & \(0.253 \pm 0.189\) & \(0.471 \pm 0.197\) & \(54.7 \pm 44.6\) & \(0.126 \pm 0.100\) \\
       Gemini 3 Flash   & \(0.962 \pm 0.096\) & \(0.501 \pm 0.234\) & \(0.818 \pm 0.164\) & \(0.255 \pm 0.210\) & \(0.469 \pm 0.218\) & \(50.2 \pm 36.4\) & \(0.119 \pm 0.098\) \\
       DeepSeek R1      & \(0.955 \pm 0.086\) & \(0.504 \pm 0.204\) & \(0.773 \pm 0.181\) & \(0.231 \pm 0.182\) & \(0.460 \pm 0.186\) & \(50.9 \pm 36.0\) & \(0.122 \pm 0.100\) \\
       QwQ 32B          & \(0.944 \pm 0.088\) & \(0.481 \pm 0.209\) & \(0.730 \pm 0.221\) & \(0.209 \pm 0.167\) & \(0.432 \pm 0.183\) & \(51.0 \pm 37.1\) & \(0.117 \pm 0.090\) \\
       Qwen 2.5 32B         & \(0.925 \pm 0.109\) & \(0.431 \pm 0.219\) & \(0.686 \pm 0.222\) & \(0.180 \pm 0.161\) & \(0.386 \pm 0.192\) & \(49.9 \pm 31.6\) & \(0.107 \pm 0.085\) \\
       Llama 3.1 8B     & \(0.916 \pm 0.156\) & \(0.341 \pm 0.205\) & \(0.615 \pm 0.256\) & \(0.111 \pm 0.130\) & \(0.295 \pm 0.171\) & \(51.8 \pm 30.9\) & \(\mathbf{0.103 \pm 0.080}\) \\
       \bottomrule
   \end{tabular}}
   \caption{Comparison of different models' performance on {\projectname} (mean $\pm$ standard deviation).}
   \vspace{-12pt}
   \label{tab:overall-results}
\end{table*}

Evaluating causal graph construction from text poses challenges that traditional metrics cannot address. First, unlike traditional causal discovery, where nodes are provided, this task requires assessing both node identification and edge prediction. Second, a generated element may be valid if supported by either the ground-truth graph or the source text; traditional metrics only compare against ground-truth graphs. Standard causal discovery metrics (precision, recall, F$_1$, Structural Hamming Distance (SHD) \cite{tsamardinos2006maxmin}) are therefore insufficient.

To address these challenges, we design an LLM-as-a-Judge framework \cite{zheng2023judgingllmasajudgemtbenchchatbot}. The framework evaluates at both node and edge levels, and compares generated elements against both the ground-truth graph and source text. Because real-world texts use synonyms and varying levels of abstraction, predicted and ground-truth elements often differ in surface form while representing the same underlying concept. The framework therefore assesses semantic similarity and abstraction levels to enable meaningful comparison. The judge processes each evaluation dimension (node precision, node recall, edge precision, edge recall) in separate prompts; combining all evaluations into a single prompt degrades accuracy due to the complexity of jointly assessing lengthy texts, generated graphs, and ground-truth graphs. We use this framework to compute three metrics:

\noindent\textbf{Precision} measures whether each generated element is valid. Each predicted node or edge is compared against both the ground-truth graph and the source text along three dimensions: (1) \textit{presence}---whether the concept appears in the reference (strong match, weak match, or no match); (2) \textit{semantic similarity}---how closely the meaning aligns (strong, moderate, weak, or N/A); and (3) \textit{abstraction level}---whether it is broader, aligned, or narrower than the reference. If an element receives ``no match'' against both ground-truth graph and source text, it scores 0.0; otherwise, the higher of the two composite scores is used.

\noindent\textbf{Recall} measures whether the generated graph captures ground-truth elements. Not all causal relationships are equally central to a graph's structure; missing a primary causal driver is more severe than missing a peripheral modifier. We therefore assess each ground-truth element for its \textit{importance} (core, intermediate, or peripheral) and how well it is captured by the generated graph. Recall is computed as a weighted average where correctness scores are multiplied by importance weights.

\noindent\textbf{SHD} is computed from the judge's edge-level evaluations as the sum of three error types: false positives (predicted edges with no ground-truth match), false negatives (ground-truth edges not captured), and reversals (edges with correct nodes but inverted direction). Normalized SHD divides by $n \times (n-1)$, where $n$ is the number of nodes, yielding a value in $[0,1]$. A comprehensive rubric and scoring mechanics are provided in Appendices \ref{app:llm-judge-scoring} and \ref{app:llm-as-a-judge-prompting}.

We validate judge reliability through cross-judge agreement analysis across four candidate judges (DeepSeek R1, DeepSeek v3.2, Gemini 3 Flash, GLM 4.7), finding strong pairwise correlations ($r = 0.78$--$0.91$) that indicate consistent rankings regardless of judge choice. We further confirm no family bias: DeepSeek v3.2 scores DeepSeek R1 outputs no more favorably than outputs from other model families (Appendix~\ref{sec:appendix-inter-judge}).

\subsection{Performance of State-of-the-Art LLMs}
\label{sec:main-results}

To assess the performance of LLMs on causal discovery from real-world text, we benchmark them on {\projectname}. As shown in Table~\ref{tab:overall-results}, all models perform poorly, with the best-performing model, Claude Opus 4.5, achieving just an average F$_1$ of $0.535$ across all samples. This shows even SOTA LLMs struggle with causal reasoning from real-world text. We observe that reasoning models show the best performance, and that there is a positive trend between model size and performance. Comparing across metrics, models perform in roughly the same ranking, showing broad agreement for overall performance. Raw SHD scores are high, as expected, due to the large size of graphs. However, normalized SHD is low, due to most graphs being sparse. Larger models and reasoning models appear to do notably better on this task. Notably, all models exhibit significantly lower recall than precision, showing that models have an easier time generating nodes and edges that are valid from the source text, but not the same as in the ground-truth graph. Fine-grained error analysis reveals that direction reversals are rare ($< 1.1\%$ of matched edges), indicating that when models identify correct relationships, they almost never invert causality. However, while 85--90\% of generated edges have some textual support, only 17--33\% match ground-truth edges, suggesting models generate plausible causal relationships from the text but fail to recover the specific relationships encoded in the ground-truth graph (Appendix~\ref{app:error_analysis}). Due to the realistic nature of the benchmark, there is a high standard deviation across metrics, with difficulty varying based on factors including explicitness, number of nodes, number of edges, text size, and domains mimicking the diverse conditions for real-world causal reasoning.

\begin{table}[h]
  \centering
  \footnotesize
  \resizebox{\columnwidth}{!}{%
  \begin{tabular}{lcccc}
      \toprule
      \textbf{Model} & \textbf{Precision~($\uparrow$)} & \textbf{Recall~($\uparrow$)} & \textbf{F$_1$~($\uparrow$)} & \textbf{SHD~($\downarrow$)} \\
      \midrule
      Claude Opus 4.5  & \textbf{0.582} & \textbf{0.536} & \textbf{0.551} & \textbf{35.6} \\
      Gemini 3 Pro    & 0.567 & 0.531 & 0.538 & 35.8 \\
      Gemini 3 Flash  & 0.552 & 0.526 & 0.528 & 37.8 \\
      GPT 5.2         & 0.537 & 0.521 & 0.520 & 42.6 \\
      Kimi K2         & 0.515 & 0.513 & 0.505 & 45.2 \\
      GLM 4.7         & 0.517 & 0.514 & 0.505 & 41.0 \\
      DeepSeek R1     & 0.513 & 0.512 & 0.502 & 40.6 \\
      QwQ 32B         & 0.483 & 0.501 & 0.481 & 41.0 \\
      Qwen 2.5 32B        & 0.290 & 0.435 & 0.332 & 51.0 \\
      Llama 3.1 8B    & 0.218 & 0.409 & 0.266 & 62.9 \\
      \bottomrule
  \end{tabular}%
  }
  \caption{Name-assisted performance where models are provided ground-truth node names. Performance remains poor despite removing node identification, showing that causal reasoning is the primary challenge.}
  \vspace{-12pt}
  \label{tab:name-assisted-results}
\end{table}

\paragraph{Node Identification vs. Causal Reasoning.}
\label{sec:ablation-name-assisted}

To investigate whether poor model performance stems from node identification or causal reasoning, and to validate our LLM-as-a-Judge framework, we conduct an ablation experiment in which models are explicitly provided with the complete set of ground-truth node names. This enables deterministic evaluation, allowing us to compare rankings across evaluation conditions.

Table~\ref{tab:name-assisted-results} shows results from this controlled scenario. While models trivially achieved perfect node-level precision and recall by design, improvements in causal inference were limited: edge-level F$_1$ slightly improved for stronger models such as DeepSeek R1 (by only $+0.042$) and Gemini 3 Pro (by $+0.037$). Meanwhile, weaker models like Qwen 2.5 32B and Llama 3.1 8B saw reductions in performance, declining by $-0.054$ and $-0.029$ respectively. SHD similarly showed minor improvements, suggesting that explicitly provided node schemas do not have a consistent effect on performance.

Though deterministic and LLM-as-a-Judge evaluations are not directly comparable, both approaches show similar results. Despite removing entity recognition uncertainty, models continue to perform poorly at correctly inferring causal relationships. This ablation highlights that the poor performance of LLMs on {\projectname} is due to fundamental limitations with causal reasoning, not entity recognition. The consistency of model rankings across deterministic and LLM-as-a-Judge evaluations provides empirical validation of our evaluation framework: systematic judge biases would manifest as ranking instability between settings, which we do not observe. We further validate robustness by computing unweighted metrics using binary matching without importance weights; model rankings remain highly consistent (Spearman's $\rho = 0.915$, Kendall's $\tau = 0.778$; Appendix~\ref{app:unweighted}).

\subsection{Performance Analysis}
\label{sec:performance-analysis}

In this section, we analyze how factors contributing to real-world complexity (explicitness, text length, number of nodes, number of edges, and domain) affect LLM causal reasoning performance.

\paragraph{Explicitness}
\label{sec:ablation-explicitness}

\begin{figure}[htbp]
  \centering
  \includegraphics[width=\columnwidth]{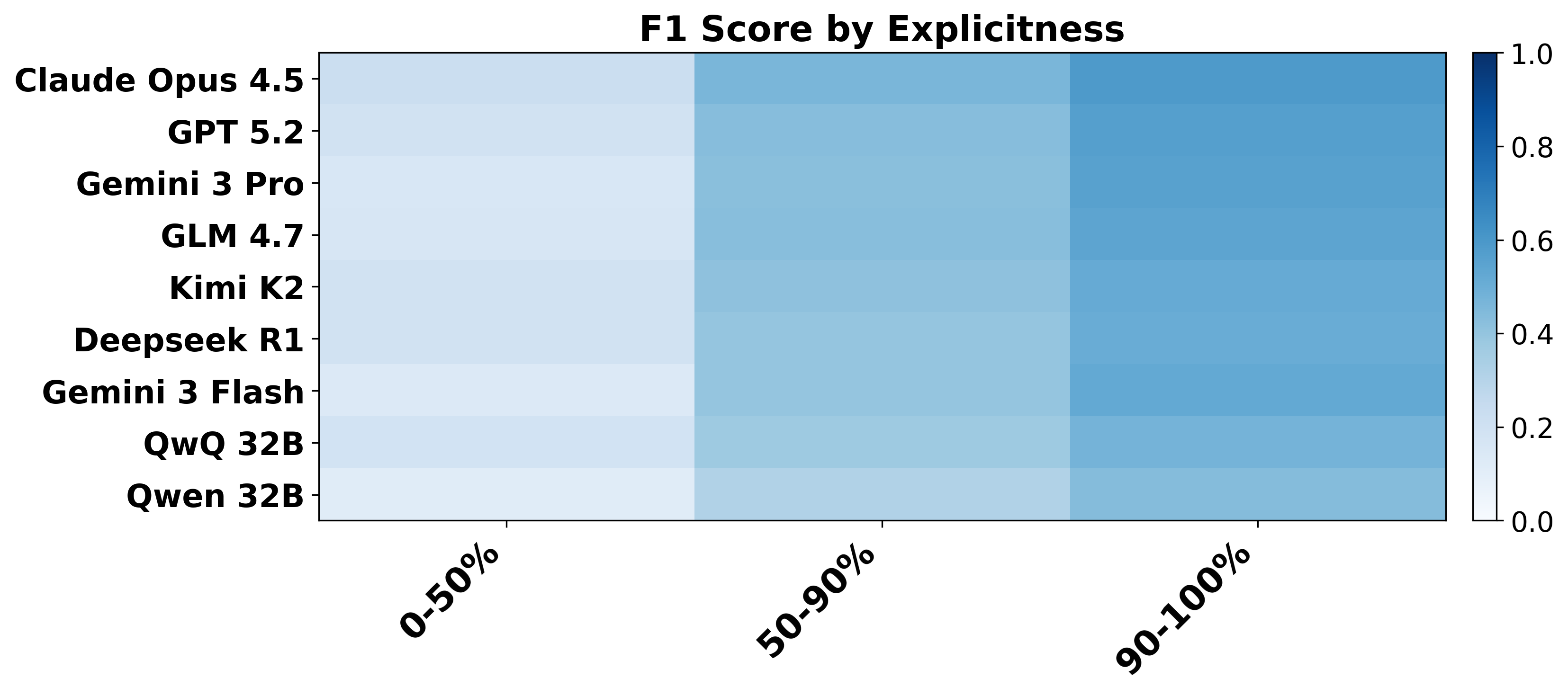}
  \caption{{Heatmap of average model F$_1$ scores across explicitness bins, where 100\% means every node is either explicitly or implicitly mentioned in the text. This shows explicitness has a large impact on performance, and LLMs struggle to infer causality when explicit references are sparse.}}
  \vspace{-12pt}
  \label{fig:explicitness-heatmap}
\end{figure}

A key challenge in real-world causal reasoning is that causal events are not always explicitly stated in the text. To analyze how this affects performance, we measure the explicitness of each sample: the degree to which causal events in the ground-truth graph are mentioned in the source text. We develop an explicitness score that categorizes each node into three levels: (1) \textit{explicit}: the node name or a clear synonym appears directly in the text, (2) \textit{implicit}: the node concept is mentioned indirectly or can be reasonably inferred, and (3) \textit{absent}: the node does not appear in the text whatsoever. We use R1 to automatically label each node in every sample using a detailed rubric (see Appendix~\ref{sec:appendix-explicitness-prompt}), then calculate the explicitness as:
\[
\mathrm{Explicitness} = \frac{1}{|V|} \sum_{v \in V}
    \begin{cases}
        1, & \text{if } v \in E \cup I,\\
        0, & \text{if } v \in A
    \end{cases}
\]

where $E$ represents explicitly mentioned nodes, $I$ represents implicitly mentioned nodes, $A$ represents nodes absent from the text, and $V$ is the set of all nodes. This acts as a natural measure of difficulty, as explicitly or implicitly mentioned nodes are easier to identify than those that are absent from the text. The explicitness score provides a quantitative way to assess how much causal reasoning (versus simple text comprehension) is required for each sample.

Samples with high degrees of explicitness are rare in a realistic benchmark. Therefore, we divide samples into three bins to ensure that each is sufficiently large to show the effect of explicitness. As shown in Figure~\ref{fig:explicitness-heatmap}, there is a strong positive correlation between a sample's explicitness and model performance (reported using F$_1$). Additionally, all models struggle even for the most explicit samples, showing that LLMs struggle with the task even under easier conditions. This performance further degrades for the least explicit texts, with F$_1$ dropping by around half for all models between the samples with lowest and highest levels of explicitness. We verify that this relationship holds under a stricter definition of explicitness in Appendix~\ref{sec:appendix-strict-explicitness}.

\paragraph{Effect of Size.}
\label{sec:ablation-effect-of-size}

\begin{table}[]
    \centering
    \small
    \setlength{\tabcolsep}{3pt}  % reduce column padding
    \begin{tabular}{lccc}
        \toprule
        \textbf{Model}    & \textbf{Text} & \textbf{Edges} & \textbf{Nodes} \\
        \midrule
        Gemini 3 Pro    & +0.08 ± 0.22 & –0.03 ± 0.22 & –0.02 ± 0.22 \\
        Claude Opus 4.5 & +0.02 ± 0.19 & –0.02 ± 0.21 & –0.00 ± 0.19 \\
        GLM 4.7         & +0.02 ± 0.20 & –0.06 ± 0.21 & –0.04 ± 0.20 \\
        Kimi K2         & +0.01 ± 0.19 & –0.03 ± 0.20 & –0.01 ± 0.20 \\
        GPT 5.2         & –0.01 ± 0.20 & –0.03 ± 0.21 & –0.02 ± 0.20 \\
        Gemini 3 Flash  & –0.01 ± 0.22 & –0.05 ± 0.22 & –0.04 ± 0.21 \\
        DeepSeek R1     & –0.01 ± 0.17 & –0.04 ± 0.19 & –0.02 ± 0.18 \\
        Qwen 2.5 32B        & –0.02 ± 0.19 & –0.12 ± 0.20 & –0.12 ± 0.19 \\
        QwQ 32B         & –0.02 ± 0.18 & –0.03 ± 0.19 & –0.03 ± 0.19 \\
        Llama 3.1 8B    & –0.03 ± 0.17 & –0.08 ± 0.18 & –0.07 ± 0.17 \\
        \bottomrule
    \end{tabular}
    \caption{Difference in model performance (mean ± std) between the top and bottom quartiles of text length, edge count, and node count. Positive values indicate improved performance on larger instances.}
    \vspace{-12pt}
    \label{tab:size-performance-delta}
\end{table}

As {\projectname} is a realistic benchmark, samples are diverse with respect to size of source text and ground-truth graph. We organize samples into quartiles for length of source text, number of edges, and number of nodes. We report the difference in performance between the small samples and large samples in Table~\ref{tab:size-performance-delta}. Counterintuitively, there is a weak, but positive correlation between size and performance. Given the strong effect of explicitness on performance shown in Section~\ref{sec:ablation-explicitness}, we investigate the relationship between explicitness and size. As samples greatly vary in size, we analyze each of them using a log scale. We find that text length, number of nodes, and number of edges each have a small, positive correlation with level of explicitness (with $R^2$ values of 0.171, 0.002, and 0.002 respectively). We find weak but significant negative correlations between graph complexity metrics (edges, motif counts) and model performance ($|r| = 0.13$--$0.18$, $p < 0.05$). However, when we bin samples by graph density or cyclicity, performance remains largely flat across both dimensions. Notably, the 90.4\% of graphs containing feedback cycles show nearly identical performance to directed acyclic graphs (Appendix~\ref{app:graph-diversity}).

\begin{figure}[ht]
  \centering
  \includegraphics[width=\columnwidth]{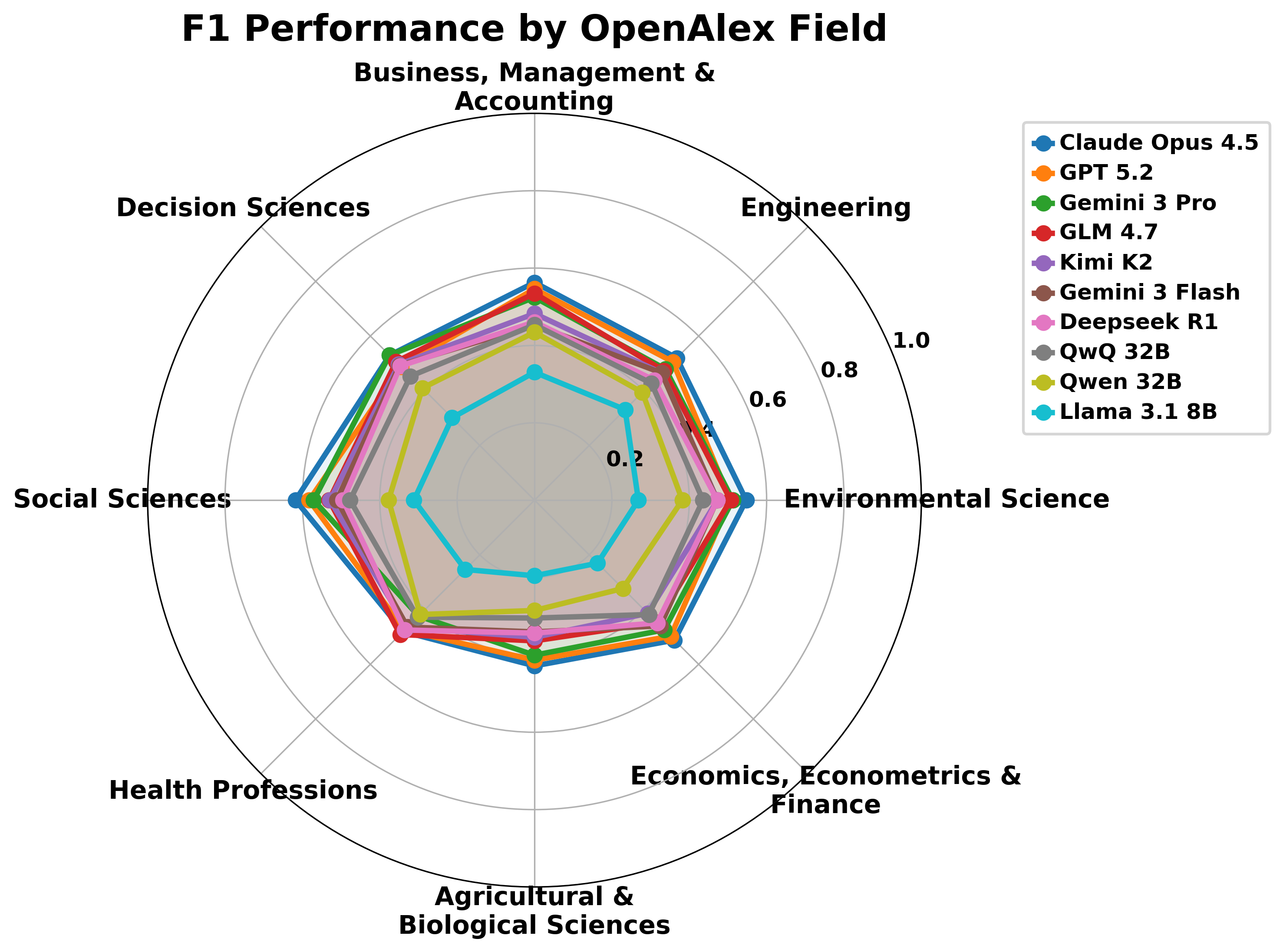}
  \caption{Radar chart of domain-specific $F_1$ scores.}
  \vspace{-12pt}
  \label{fig:radar-domain-accuracies}
\end{figure}

\paragraph{Effect of Domain and Data Diversity.}
\label{sec:domain-diversity}

To assess whether semantic domain affects performance, we cluster paper embeddings using k-means (k=4, selected by silhouette score) and analyze F$_1$ by cluster. HDBSCAN found no natural clusters, classifying all papers as noise. Cluster differences are not statistically significant (ANOVA $F = 1.03$, $p = 0.38$, $\eta^2 = 0.012$; Kruskal-Wallis $H = 3.42$, $p = 0.33$), with a performance gap of only $0.041$ $F_1$ between best and worst clusters. As shown in Figure~\ref{fig:radar-domain-accuracies}, the best-performing model overall, Claude Opus 4.5, outperforms on 6 out of 8 domains, while Gemini 3 Pro has a higher score in Decision Sciences and GLM 4.7 in Health Professions. However, all models perform poorly in all domains, indicating that domain does not substantially impact difficulty. We also analyze semantic diversity via embedding-based comparisons against reference corpora and test whether semantic domain predicts task difficulty through cluster-based performance analysis, as reported in Appendix~\ref{app:textual-diversity}.

\subsection{Case Study}
\label{sec:case-study-r1-response}

\begin{figure}[t]
  \centering
  \begin{tcolorbox}[colback=blue!10, colframe=blue, title=R1 Chain-of-Thought Excerpt]
  …In the Materials and Methods, they mention… auxiliary variables (savings, non-farm income, \textbf{per capita arable land}), and constants (region area, \textbf{precipitation}, etc.)...\\
  County's arable land area (18) -$>$ Cash crop cultivation (10) (if less land, shift to cash crops)... Cash crop cultivation (10) -$>$ Farmers' income (maybe part of non-farm income, but non-farm income is separate. Cash crop would be part of farm income...\\
  county financial capital increases lead to infrastructure and loan access...\\
  - Loan usage flexibility -$>$ Diversified investments (cash crops, etc.)
  \end{tcolorbox}
  \caption{Verbatim excerpts from R1's reasoning trace. Ellipses added for readability and bold for emphasis.}
  \vspace{-12pt}
  \label{fig:cot-excerpt}
\end{figure}

To further analyze {\projectname}, we select a benchmark sample about livelihood efficiency in the Qinba Mountains and the graph constructed by R1. For illustrative purposes, we focus on a subset of the graph showing the causal relationship between various climate and land factors with grain production and overall output. The source text describes these relationships most clearly in this passage:
\begin{quote}
``Land and climate are the fundamental conditions for agricultural production. … therefore, sunshine, precipitation, and arable land area were selected to represent the natural capital of the county.''
\end{quote}

\begin{figure}[t]
  \centering
  \resizebox{\columnwidth}{!}{%
    \begin{tikzpicture}[
        >={Stealth[round,length=8pt,width=6pt]},
        capturedvar/.style={
          circle, draw=green!60!black, fill=green!15, minimum size=2.5cm,
          font=\small\bfseries, align=center, inner sep=3pt
        },
        missedvar/.style={
          circle, draw=blue!60!black, fill=blue!15, minimum size=2.5cm,
          font=\small\bfseries, align=center, inner sep=3pt
        },
        spuriousvar/.style={
          circle, draw=red!60!black, fill=red!15, minimum size=2.2cm,
          font=\small\bfseries, align=center, inner sep=3pt
        },
        missededge/.style={->, very thick, black, dashed},
        spuriousedge/.style={->, very thick, red}
      ]
      %--- Ground-truth graph with R1 performance annotations ---
      \node[capturedvar] (land)   at (-4.5, -2.0) {Arable land\\in county};
      \node[missedvar]   (precip) at (-4.5,  2.0) {Annual\\precipitation};
      \node[missedvar]   (sun)    at (-4.5,  0.0) {Annual\\sunshine hours};
      \node[missedvar]   (grain)  at ( 0.5,  0.0) {Total grain\\output};
      
      \draw[missededge] (precip) -- (grain);
      \draw[missededge] (sun)    -- (grain);
      \draw[missededge] (land)   -- (grain);
      
      %--- Spurious additions by R1 ---
      \node[spuriousvar] (loans) at ( 3.5,  1.5) {Loan usage\\flexibility};
      \node[spuriousvar] (crops) at ( 3.5, -2.0) {Cash crop\\cultivation};
      \draw[spuriousedge] (land) -- (crops);
      \draw[spuriousedge] (loans) -- (crops);
    \end{tikzpicture}%
  }
  \caption{Causal subgraph annotated with R1's performance. Green: concepts R1 correctly identified. Blue: ground-truth nodes not included. Dashed: ground-truth edges not included. Red: spurious nodes and edges.}
  \vspace{-12pt}
  \label{fig:combined-slices}
\end{figure}

The ground-truth graph captures these relationships directly: \emph{Annual precipitation}, \emph{Annual sunshine hours}, and \emph{Arable land in county} each cause \emph{Total grain output}. Despite this being explicitly described in the source text, R1 fails to faithfully reconstruct the graph. As shown in Figure~\ref{fig:cot-excerpt}, R1 initially identifies precipitation as a relevant factor during reasoning, but omits both \emph{Annual precipitation} and \emph{Annual sunshine hours} from its final graph. While R1 correctly identifies \emph{Arable land in county}, it connects this node to \emph{Cash crop cultivation} rather than \emph{Total grain output}. R1 also introduces spurious structure, adding \emph{Loan usage flexibility} linked to \emph{Cash crop cultivation}, despite loan flexibility relating to financial capital, not crop selection. As shown in Figure~\ref{fig:combined-slices}, R1 recovers only 1 of 4 ground-truth nodes, misses all 3 ground-truth edges, and adds 2 spurious nodes with 2 spurious edges. This illustrates how LLMs struggle to integrate causally relevant information dispersed across lengthy real-world documents, even when individual cues are straightforward. We include a complementary human expert case study in Appendix~\ref{app:expert-subgraph}, demonstrating that the task is tractable.

\section{Conclusions}
\label{sec:conclusions}

In this paper, we introduce {\projectname}, the first benchmark to assess causal reasoning from text under realistic conditions. {\projectname} draws diverse samples from academic literature, featuring texts varying in length, graph complexity, and domain. Experiments show that SOTA LLMs struggle significantly, with the best model achieving just 0.535 F$_1$. Performance degrades further as causal relationships become less explicit, and remains poor even when node names are provided, indicating the primary challenge is causal reasoning itself, not entity recognition. {\projectname} provides a cost-effective platform (all experiments under \$1000) for future research into these fundamental limitations.

\section*{Limitations}
\label{sec:limitations}

We acknowledge the following limitations: (i) Our evaluation methodology incorporates LLMs for post-processing and judging, which may introduce model-specific biases and calibration differences. We validate reliability through cross-judge agreement analysis (showing strong pairwise correlations, $r > 0.78$) and verify absence of family bias, though LLM judgments may not perfectly align with expert human preferences. (ii) Models are provided with the target number of nodes, simplifying the task compared to fully unconstrained causal discovery; this constraint is necessary to disentangle causal reasoning ability from granularity preferences and to enable consistent evaluation across samples with varying complexity. (iii) Our corpus is drawn from open-access academic literature, and while we demonstrate diversity across domains and graph structures, performance on {\projectname} may not fully generalize to other text genres or domains not represented in the benchmark. (iv) Our benchmark is English-only; while the underlying causal reasoning task is language-agnostic, performance on {\projectname} may not generalize to other languages. (v) We evaluate models exclusively in the zero-shot setting with default hyperparameters. While all models have been trained to utilize chain-of-thought (which they were prompted to do), techniques like few-shot prompting or retrieval-augmented generation may further improve performance. (vi) {\projectname} draws from academic papers employing causal loop diagrams, a methodology from the field of system dynamics. Model performance on this benchmark may not generalize to causal graphs constructed under different paradigms.

\section*{Ethics Statement}
\label{sec:ethics-statement}

{\projectname} is constructed from open-access academic papers published under CC BY 4.0 licenses (MDPI and PLOS), which permit redistribution and derivative works. We do not anticipate significant risk of personally identifiable information or offensive content, as the source materials are academic publications; we did not perform additional screening for such content beyond the peer review already applied by the source venues. Annotation was performed by undergraduate economics students, whose participation was optional and who were informed that their annotations would be used in an academic publication. Further details are provided in Appendix~\ref{app:inter-annotator}. {\projectname} is intended for research evaluation of LLMs to better understand their causal reasoning capabilities. We acknowledge there is always a risk of misuse by malicious actors, but believe this risk is outweighed by furthering the understanding of these models' capabilities. LLMs were heavily used in this work for the data processing pipeline, as detailed in Section~\ref{sec:dataset-creation}. They were also used to aid with writing code, literature review, and minor revisions to the paper; however, all AI-related changes were manually reviewed and approved by the authors.

\bibliography{custom}

\appendix

\begin{table*}[t!]
    \centering
    \resizebox{\textwidth}{!}{%
    % Column spec: l l l l c c c c (8 columns)
    \begin{tabular}{@{}llllllll@{}}
        \textbf{Benchmark} & \textbf{Primary Task} & \textbf{Long Text} & \textbf{Input Type} & \textbf{Max Nodes} & \textbf{Diverse} & \textbf{Complex} & \textbf{Realism} \\
        \toprule
        % Group 1: Pairwise or Highly Explicit/Controlled
        \textbf{ExpliCa \cite{miliani2025explicaevaluatingexplicitcausal}} & Pairwise ID & \xmark & Sentences & 2 & \xmark & \xmark & \xmark \\
        \textbf{LLM Fallacies \cite{joshi2024llmspronefallaciescausal}} & Pairwise Inference & \xmark & Scenario/Vignette & 2 & \xmark & \partialmarktext & \xmark \\
        \textbf{Plausibly Exogenous \cite{Oh2025}}  & Pairwise ID & \cmark & Full Document & 2 & \cmark & \partialmarktext & \cmark \\
        \textbf{Lee et al. \cite{lee2025benchmarkingllmcausalreasoning}} & Question About Triplet & \xmark & Triplet/Question & 3 & \cmark & \xmark & \xmark \\
        
        % Group 2: Graph Construction from Short/Simplified Descriptions
        \textbf{From text to map \cite{hosseinichimeh2024}} & Graph Construction & \xmark & Short Narratives & 15 & \xmark & \xmark & \cmark \\
        \textbf{Failure Modes \cite{yamin2024failuremodesllmscausal}} & ID/Graph Construction & \xmark & Short Narrative & 20 & \xmark & \partialmarktext & \partialmarktext \\
        
        % Group 3: Sentence-Level Analysis or Broader Surveys
        \textbf{From Text to Model \cite{Veldhuis2024}} & Sentence Classification & \xmark & Sentences & N/A & \xmark & \partialmarktext & \cmark \\
        \textbf{Causal Reasoning Survey \cite{kıcıman2024causalreasoninglargelanguage}} & Multiple Tasks & \xmark & Mixed & Varies & \partialmarktext & \partialmarktext & \partialmarktext \\
        \textbf{CausalTalk \cite{ding-etal-2025-multi}} & Multi-task Causal Language & \xmark & Social Media Posts & N/A & \cmark & \partialmarktext & \cmark \\
    
        % Group 4: QA or Structure ID from More Structured/Abstracted Data
        \textbf{CaLM \cite{chen2024causalevaluationlanguagemodels}} & Multiple Tasks & \xmark & Mixed & 20 & \partialmarktext & \xmark & \xmark \\
        \textbf{CausalProbe-2024 \cite{chi2025unveilingcausalreasoninglarge}} & Causal QA & \xmark & QA/Short Context & N/A & \xmark & \partialmarktext & \partialmarktext \\
        \textbf{CLadder \cite{Jin2024}} & Causal QA & \xmark & Narrative/Question & 4 & \xmark & \cmark & \xmark \\
        \textbf{CausalGraphBench \cite{babakov-etal-2025-causalgraphbench}} & Causal Structure ID & \xmark & Variable List & 222 & \partialmarktext & \partialmarktext & \xmark \\
        \textbf{CausalBench \cite{zhou2024causalbenchcomprehensivebenchmarkcausal}} & Causal Structure ID & \xmark & Query/Question & 109 & \partialmarktext & \partialmarktext & \cmark \\
        \textbf{COLD \cite{joshi2024coldcausalreasoningclosed}} & Causal QA & \xmark & Query/Question & 33 & \partialmarktext & \cmark & \cmark \\
        \textbf{CaLQuest.PT \cite{lasheras-pinheiro-2025-calquest}} & Causal QA & \xmark & Query/Question & N/A & \cmark & \cmark & \cmark \\
        \midrule
        \textbf{{\projectname} (Ours)} & Graph Construction & \cmark & Full Document & 140 & \cmark & \cmark & \cmark \\
        \bottomrule
    \end{tabular}%
    }
    \caption{Visual comparison of {\projectname} with other causal reasoning benchmarks.
    \textbf{Primary Task} (e.g., Graph Construction: Graph Construction; Pairwise ID: Pairwise Identification; QA: Question Answering).
    \textbf{Long Text}: Indicates if the benchmark primarily uses long textual inputs.
    \textbf{Input Type} (e.g., Document, Narrative, Scenarios, Queries, Sentences).
    \textbf{Max Nodes}: Maximum nodes per instance for graph construction or the underlying model.
    Symbols for realism criteria (\textbf{Diverse}, \textbf{Complex}, \textbf{Realism}): \cmark: Fully meets criterion; \partialmarktext: Partially meets criterion; \xmark: Does not substantially meet criterion, relative to {\projectname}'s focus on long-text graph extraction.}
    \label{tab:appendix-benchmark-comparison}
\end{table*}

\section{Detailed Comparison with Causal Reasoning Benchmarks}
\label{sec:detailed-comparison}
This section provides a comparative overview of {\projectname} against existing benchmarks for causal reasoning. While each benchmark contributes to understanding LLM capabilities, {\projectname} is specifically designed to test the construction of complex causal graphs from long-form, real-world academic texts. We evaluate benchmarks based on their diversity and complexity. In this context, we consider a benchmark diverse when it spans many domains or sub-domains, or draws data from many different types of sources. Meanwhile, we deem a sample complex when it exhibits varied levels of explicitness; that is, when the benchmark contains many samples that are not highly explicit and therefore require causal reasoning rather than mere reading comprehension. Lastly, we examine the realism of the source data: is it drawn from real-world settings? The following table offers a high-level visual comparison, with further details on each benchmark discussed subsequently.

\subsection{Discussions of Selected Benchmarks}
The following discussions provide context for the data presented in Table~\ref{tab:appendix-benchmark-comparison}, highlighting their approaches and how our benchmark differs from prior work. While these works are valuable, prior work fails to measure the causal reasoning abilities of LLMs from text under real-world settings. We highlight the approach of each benchmark, and how they compare to {\projectname}.

Several benchmarks concentrate on pairwise causal relations, include inputs that are highly explicit, or have inputs that are generated synthetically or are crafted as short texts by hand. We differ from these approaches by aiming to construct large graphs from real-world literature. \textbf{ExpliCa \cite{miliani2025explicaevaluatingexplicitcausal}} examines how LLMs understand explicit connectives in sentence pairs, resulting in 2-node links. By design, it uses very short, often crafted inputs and focuses on explicit cues, thereby avoiding the complexities of implicit causality and information integration from extensive texts that {\projectname} targets. Meanwhile, \textbf{LLM Fallacies \cite{joshi2024llmspronefallaciescausal}} employs short, synthetic scenarios to test pairwise causal inference, focusing on logical fallacies when LLMs are presented with explicit non-causal information. While this addresses a specific type of reasoning complexity, its synthetic and brief inputs differ greatly from the real-world, extensive texts and broader graph construction task in {\projectname}. \textbf{Plausibly Exogenous Galore \cite{Oh2025}} serves as an interesting bridge, as it uses long economics documents similar to {\projectname}. However, its task is to find only the main pairwise link for the entire document, which greatly limits its diversity and complexity.

Another group of benchmarks attempt graph construction, but typically rely on short, simplified, or synthetic descriptions. These lack the depth and realism of long, real-world texts. \textbf{From text to map \cite{hosseinichimeh2024}} generates relatively small graphs (max 9--15 nodes) from concise, hand-crafted descriptions. Such inputs inherently limit textual diversity and likely feature more explicit causal links, sidestepping the challenge of parsing lengthy, nuanced documents with varying levels of explicitness. \textbf{Failure Modes of LLMs for Causal Reasoning on Narratives \cite{yamin2024failuremodesllmscausal}} also uses short, often synthetic or CauseNet-derived narratives for constructing linear chain graphs (max 20 nodes). While it explores LLM biases and indirect effects, its input lacks the textual diversity and structural graph complexity of {\projectname}, and its narratives are purpose-built rather than reflecting reasoning under real-world conditions. Lee et al. \cite{lee2025benchmarkingllmcausalreasoning} extract causal triplets from titles and abstracts of literature, and ask questions about that ground-truth triplet. However, unlike {\projectname}, this relies on the model's pre-existing knowledge as no source text is provided as grounding, and tests only for the ability to answer binary questions rather than open-ended causality. Similar to {\projectname}, \textbf{CausalGraphBench} \cite{babakov-etal-2025-causalgraphbench} derives its samples from published causal graphs; however, its task is not causal reasoning from text, requiring only determining the connections between the provided list of nodes given a short synthetic description (e.g., a 78 word description for a graph with over 750 edges).

Other benchmarks focus on a sentence-level analysis, or offer broader surveys of causal tasks where individual components may use non-primary inputs or address different facets of reasoning. \textbf{From Text to Model \cite{Veldhuis2024}} (NLP for SD) measures the ability of LLMs to classify individual sentences from real-world texts for causality. While it uses real-world text, it uses small excerpts, and avoids the complexities of reasoning over large documents. \textbf{Causal Reasoning Survey \cite{kıcıman2024causalreasoninglargelanguage}} provides a wide-ranging overview of LLM capabilities across multiple causal tasks. However, its sub-tasks often use short or structured inputs (e.g., variable lists for graph construction, concise vignettes for reasoning), which differ from {\projectname}'s reliance on extensive, unmodified academic texts for end-to-end graph extraction. CausalTalk \cite{ding-etal-2025-multi} is another form of sentence-level analysis, where evaluated models are instructed to extract the shortest phrase that describes the explicit or implicit causal relationship from a social media post. While this uses real-world text, it uses small excerpts and does not require the model to do open-ended reasoning, thus avoiding many of the complexities of real-world texts where identifying causality is not as simple as quoting it directly from the document.

To the best of our knowledge, {\projectname} is the first benchmark for LLMs to measure causal reasoning abilities from long, diverse, and complex, real-world texts. While previous benchmarking efforts have explored lengthy, diverse, and complex texts, causal reasoning, and real-world conditions separately, we are the first to do so simultaneously.

\section{Effect of Knowledge Cutoff}
\label{app:knowledge-cutoff}

\begin{table}[h]
\centering
\small
\resizebox{\columnwidth}{!}{
\begin{tabular}{lll}
\toprule
\textbf{Model} & \textbf{Release Date} & \textbf{Knowledge Cutoff} \\
\midrule
Claude Opus 4.5 & Nov.\ 2025 & May 2025 \\
GPT 5.2 & Dec.\ 2025 & Aug.\ 31, 2025 \\
Gemini 3 Pro & Nov.\ 2025 & Jan.\ 2025 \\
Gemini 3 Flash & Dec.\ 2025 & Jan.\ 2025 \\
GLM 4.7 & Dec.\ 2025 & N/A \\
Kimi K2 & Jul.\ 2025 & N/A \\
DeepSeek R1 & Jan.\ 2025 & Jul.\ 2024 \\
QwQ 32B & Mar.\ 2025 & N/A \\
Qwen 2.5 32B & Sep.\ 2024 & Sep.\ 2024 \\
Llama 3.1 8B & Jul.\ 2024 & Dec.\ 2023 \\
DeepSeek v3.2 & Dec.\ 2025 & N/A \\
\bottomrule
\end{tabular}
}
\caption{Release dates and publicly available knowledge cutoffs for evaluated models and judge models. N/A denotes an unknown knowledge cutoff date.}
\label{tab:model-recency}
\end{table}

To investigate whether performance on {\projectname} is influenced by pre-training contamination, we conduct an analysis comparing performance for each model before and after their knowledge cutoff. Table~\ref{tab:model-recency} is provided to show the knowledge cutoffs of each model utilized in {\projectname}. We select three models with varying cutoff dates: Llama 3.1 8B (December 2023), R1 (July 2024), and Qwen 2.5 32B (September 2024). Our dataset contains 35 samples derived from papers published in 2024 or later. For Llama 3.1 8B, the average F$_1$ score increased from 0.288 on samples from before its knowledge cutoff to 0.339 on newer samples; for R1, performance increased slightly from 0.459 to 0.476; while Qwen 2.5 32B showed a decrease from 0.388 to 0.342. Despite this variation in performance, none of these differences are statistically significant. This suggests that models' ability to identify causal relationships is not related to training on these specific documents during pre-training.

\section{Inter-Judge Agreement and Bias Analysis}
\label{sec:appendix-inter-judge}

To validate our LLM-as-a-Judge evaluation methodology, we measured agreement between multiple judge models and analyzed potential biases.

\subsection{Overall Inter-Judge Agreement}

Table~\ref{tab:inter-judge-agreement} reports pairwise correlations between four candidate judges on overlapping evaluation subsets.

\begin{table}[htbp]
    \centering
    \resizebox{\columnwidth}{!}{%
    \begin{tabular}{llcccc}
        \toprule
        \textbf{Judge 1} & \textbf{Judge 2} & \textbf{n} & \textbf{$r$} & \textbf{Mean $\Delta$} & \textbf{Std $\Delta$} \\
        \midrule
        DeepSeek R1   & DeepSeek v3.2  & 959 & $0.820$ & $+0.004$ & $0.120$ \\
        DeepSeek v3.2 & Gemini 3 Flash & 595 & $0.895$ & $-0.122$ & $0.090$ \\
        DeepSeek R1   & Gemini 3 Flash & 463 & $0.782$ & $-0.116$ & $0.134$ \\
        \bottomrule
    \end{tabular}%
    }
    \caption{Inter-judge agreement on large evaluation subsets. All correlations are strong ($r > 0.78$, $p < 10^{-30}$), indicating consistent rankings. Mean $\Delta$ shows differences (positive denotes first judge scores higher).}
    \label{tab:inter-judge-agreement}
\end{table}

All judge pairs show strong correlations ($r = 0.78$--$0.90$), indicating that all LLM judges produce consistent rankings regardless of differences in raw scores. This high level of agreement lends credibility to the effectiveness of our judging method.

\subsection{Cross-Judge Analysis}

To further validate consistency, we had all four candidate judges (DeepSeek R1, DeepSeek v3.2, Gemini 3 Flash, and GLM 4.7) evaluate the same 41 randomly selected DeepSeek R1 outputs. Table~\ref{tab:cross-judge} shows all pairwise correlations on this matched subset.

\begin{table}[htbp]
    \centering
    \resizebox{\columnwidth}{!}{%
    \begin{tabular}{llcc}
        \toprule
        \textbf{Judge 1} & \textbf{Judge 2} & \textbf{$r$} & \textbf{Mean $\Delta$} \\
        \midrule
        DeepSeek R1    & DeepSeek v3.2  & $0.863$ & $-0.055$ \\
        DeepSeek R1    & Gemini 3 Flash & $0.912$ & $+0.070$ \\
        DeepSeek R1    & GLM 4.7        & $0.788$ & $-0.020$ \\
        DeepSeek v3.2  & Gemini 3 Flash & $0.863$ & $-0.055$ \\
        DeepSeek v3.2  & GLM 4.7        & $0.912$ & $+0.070$ \\
        Gemini 3 Flash & GLM 4.7        & $0.912$ & $+0.070$ \\
        \bottomrule
    \end{tabular}
    }
    \caption{Pairwise judge agreement on 41 matched samples evaluated by all four judges. There is a strong correlation ($r = 0.79$--$0.91$) across all pairs, validating that inter-judge agreement is not specific to a particular judge.}
    \label{tab:cross-judge}
\end{table}

\subsection{Family Bias Analysis}

A potential concern with LLM judges is self-bias. While our judge, DeepSeek v3.2, is not an evaluated model, it is in the DeepSeek family of models, and thereby could score DeepSeek R1 more favorably. To test this, we compare the calibration gap between v3.2 and Gemini 3 Flash when judging outputs from different model families.

\begin{table}[htbp]
    \centering
    \resizebox{\columnwidth}{!}{%
    \begin{tabular}{lccc}
        \toprule
        \textbf{Outputs From} & \textbf{n} & \textbf{v3.2 -- Gemini} & \textbf{DeepSeek?} \\
        \midrule
        DeepSeek R1  & 292 & $-0.128$ & Yes \\
        Llama 3.1 8B & 214 & $-0.114$ & No \\
        QwQ 32B      &  89 & $-0.122$ & No \\
        \bottomrule
    \end{tabular}
    }
    \caption{Calibration gap between DeepSeek v3.2 and Gemini 3 Flash by output source. Negative values indicate v3.2 scores higher. The gap is consistent ($-0.11$ to $-0.13$) regardless of whether the outputs come from a DeepSeek model, suggesting no detectable family bias.}
    \label{tab:family-bias}
\end{table}

DeepSeek v3.2 scores outputs approximately 0.12 points higher than Gemini 3 Flash across all model families. The difference between judging DeepSeek R1 outputs ($-0.128$) versus non-DeepSeek outputs ($-0.114$ to $-0.122$) is minimal (0.006--0.014) and well within the noise of calibration differences. This suggests that v3.2 does not give preferential treatment to DeepSeek outputs.

\subsection{Judge Selection}

We select DeepSeek v3.2 as our primary judge because it is not among the evaluated models, performs strongly at long context lengths, shows high agreement with other judges ($r > 0.82$), and has tractable computational costs.

\section{Unweighted Evaluation Metrics}
\label{app:unweighted}
We also report an unweighted version of our fine-grained evaluation, acting as an alternative measure of performance. This metric uses binary matching (any match versus no match), removes importance weighting, partial credit for the wrong level of abstraction, and does not provide credit when the generated graph includes elements that are supported by the text but not the graph. Specifically, for unweighted evaluation, we derive binary decisions from the LLM judge's labels:
\begin{itemize}
    \item \textbf{Match}: \textbf{PRESENCE\_STRONG\_MATCH} or \textbf{PRESENCE\_WEAK\_MATCH}
    \item \textbf{No Match}: \textbf{PRESENCE\_NO\_MATCH}
\end{itemize}
Precision and recall are computed against the ground-truth graph only (with no credit for grounding solely from the text), with all elements weighted equally:
\begin{align}
    \text{Precision} &= \frac{|\text{Predicted} \cap \text{Ground-Truth}|}{|\text{Predicted}|} \\
    \text{Recall} &= \frac{|\text{Predicted} \cap \text{Ground-Truth}|}{|\text{Ground-Truth}|}
\end{align}
Tables~\ref{tab:unweighted-edges} and~\ref{tab:unweighted-nodes} present the unweighted metrics. Under this stricter evaluation, edge F$_1$ scores range from 0.111 to 0.297, substantially lower than the fine-grained scores in Table~\ref{tab:overall-results}. Additionally, the gap between node and edge performance is even more pronounced under unweighted evaluation. The best-performing model achieves 0.592 node F$_1$ but only 0.297 edge F$_1$, reinforcing our finding that causal edge inference, rather than node identification, is the primary bottleneck. Model rankings remain largely consistent with the fine-grained evaluation, with Claude Opus 4.5 leading across all metrics and Llama 3.1 8B trailing. Notably, GPT 5.2 exhibits the highest edge recall (0.290) but lower precision (0.258), suggesting it generates more edges overall at the cost of accuracy.

\paragraph{Rank Agreement.} To quantify the consistency between weighted and unweighted evaluation, we compute rank correlation statistics across overall F$_1$ scores. The two methods show strong agreement (Spearman's $\rho = 0.915$, $p < 0.001$; Kendall's $\tau = 0.778$, $p < 0.001$). The top-performing model (Claude Opus 4.5) and bottom three models (QwQ 32B, Qwen 2.5 32B, Llama 3.1 8B) maintain identical rankings across both metrics. Minor rank changes occur in the middle of the distribution: GPT 5.2 drops two positions under unweighted evaluation due to its recall-heavy strategy receiving less benefit, while DeepSeek R1 gains two positions. This strong rank correlation validates that our fine-grained scoring preserves relative model performance while providing more nuanced assessment of partial matches and abstraction differences.

\begin{table}[h]
    \centering
    \begin{tabular}{lccc}
        \toprule
        \textbf{Model} & \textbf{Prec.} & \textbf{Recall} & \textbf{F$_1$} \\
        \midrule
        Claude Opus 4.5  & \textbf{0.350} & 0.283 & \textbf{0.297} \\
        Gemini 3 Pro     & 0.317 & 0.266 & 0.275 \\
        GLM 4.7          & 0.304 & 0.252 & 0.260 \\
        GPT 5.2          & 0.258 & \textbf{0.290} & 0.253 \\
        Gemini 3 Flash   & 0.282 & 0.239 & 0.239 \\
        Kimi K2          & 0.267 & 0.237 & 0.229 \\
        DeepSeek R1      & 0.277 & 0.217 & 0.226 \\
        QwQ 32B          & 0.272 & 0.201 & 0.213 \\
        Qwen 2.5 32B         & 0.228 & 0.170 & 0.173 \\
        Llama 3.1 8B     & 0.188 & 0.105 & 0.111 \\
        \bottomrule
    \end{tabular}
    \caption{Unweighted edge-level metrics from the LLM Judge using binary matching against the ground-truth graph.}
    \label{tab:unweighted-edges}
\end{table}
\begin{table}[h]
    \centering
    \begin{tabular}{lccc}
        \toprule
        \textbf{Model} & \textbf{Prec.} & \textbf{Recall} & \textbf{F$_1$} \\
        \midrule
        Claude Opus 4.5  & \textbf{0.603} & \textbf{0.600} & \textbf{0.592} \\
        Gemini 3 Pro     & 0.568 & 0.557 & 0.551 \\
        DeepSeek R1      & 0.568 & 0.553 & 0.545 \\
        GPT 5.2          & 0.551 & 0.557 & 0.543 \\
        GLM 4.7          & 0.560 & 0.551 & 0.543 \\
        QwQ 32B          & 0.550 & 0.522 & 0.522 \\
        Kimi K2          & 0.530 & 0.543 & 0.518 \\
        Gemini 3 Flash   & 0.519 & 0.534 & 0.514 \\
        Qwen 2.5 32B        & 0.566 & 0.476 & 0.489 \\
        Llama 3.1 8B     & 0.508 & 0.384 & 0.402 \\
        \bottomrule
    \end{tabular}
    \caption{Unweighted node-level metrics from the LLM Judge using binary matching against the ground-truth graph.}
    \label{tab:unweighted-nodes}
\end{table}

\section{Graph Structural Diversity}
\label{app:graph-diversity}

\begin{figure*}[!t]
    \centering
    \includegraphics[width=0.32\textwidth]{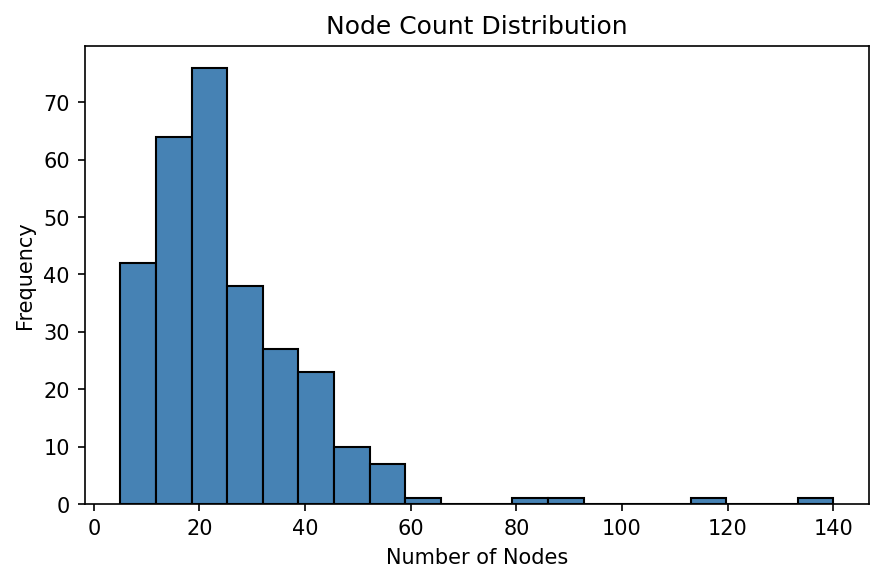}
    \hfill
    \includegraphics[width=0.32\textwidth]{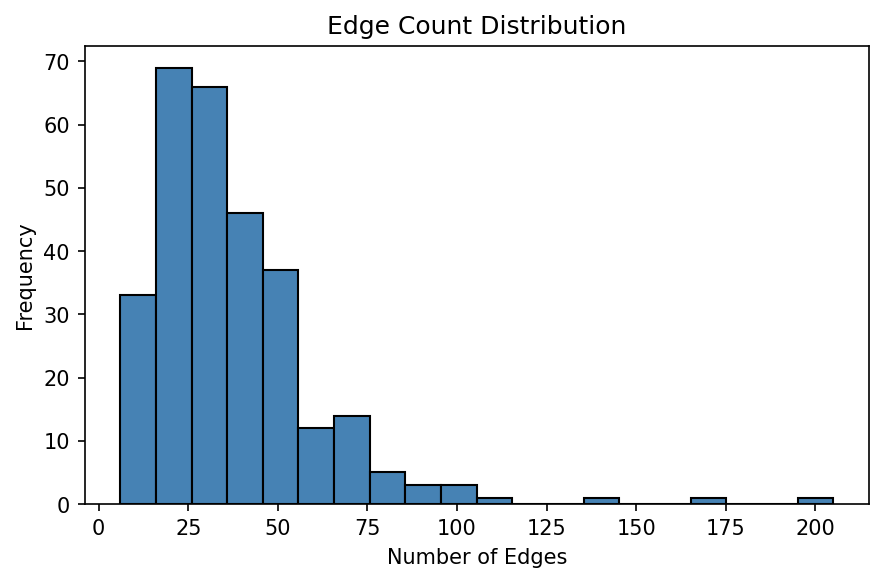}
    \hfill
    \includegraphics[width=0.32\textwidth]{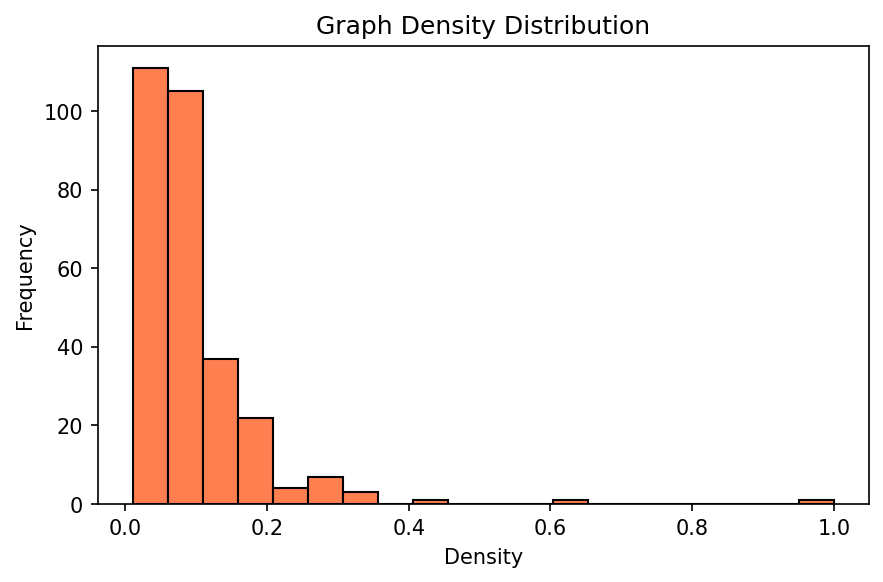}
    
    \vspace{0.2cm}
    
    \includegraphics[width=0.32\textwidth]{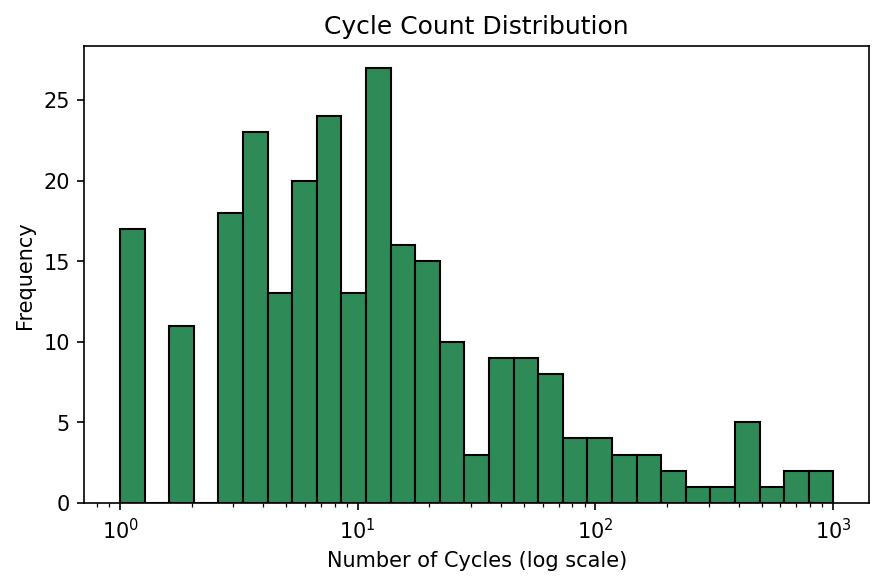}
    \hfill
    \includegraphics[width=0.32\textwidth]{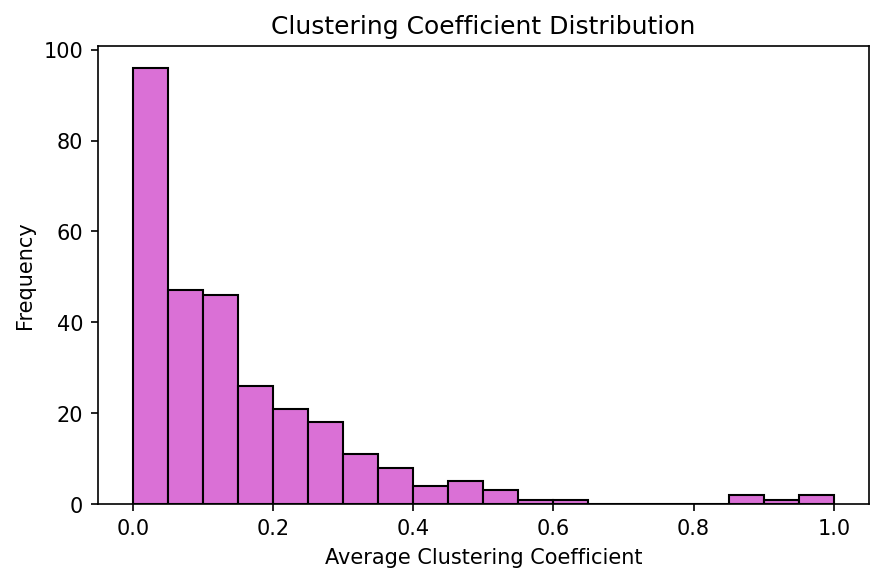}
    \hfill
    \includegraphics[width=0.32\textwidth]{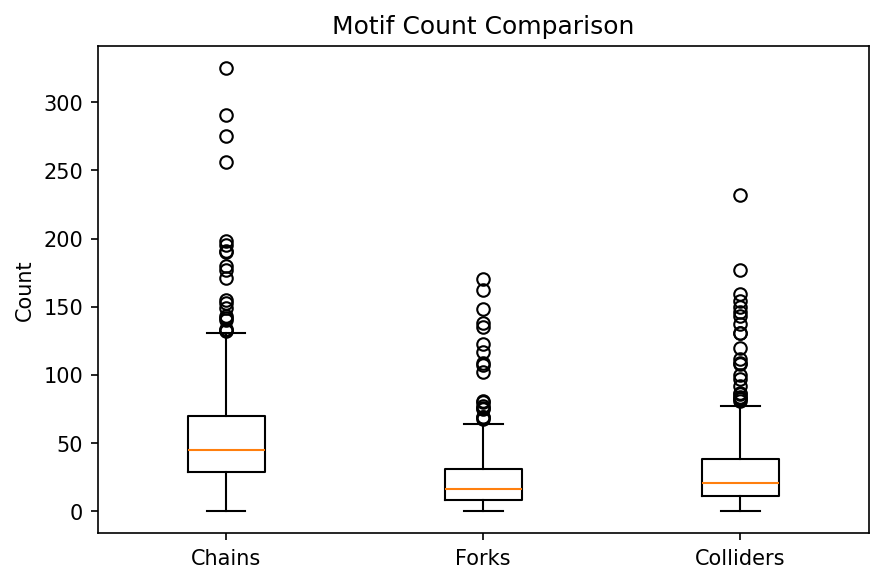}
    \caption{Distribution of graph topology metrics. Top row: node count, edge count, density. Bottom row: cycle count (log scale), clustering coefficient, motif counts.}
    \label{fig:topology-distributions}
\end{figure*}

We compute graph topology metrics across all 292 benchmark samples to characterize structural diversity. Table~\ref{tab:topology-metrics} reports summary statistics and Figure~\ref{fig:topology-distributions} visualizes distributions.

\begin{table}[h]
\centering
\resizebox{\columnwidth}{!}{
\begin{tabular}{lrrrrr}
\toprule
\textbf{Metric} & \textbf{Mean} & \textbf{Std} & \textbf{Min} & \textbf{Median} & \textbf{Max} \\
\midrule
Nodes & 25.0 & 15.8 & 5 & 22.0 & 140 \\
Edges & 37.0 & 23.9 & 6 & 32.5 & 205 \\
Density & 0.094 & 0.089 & 0.011 & 0.070 & 1.000 \\
Avg Degree & 3.0 & 0.8 & 1 & 2.9 & 10 \\
Sources & 4.8 & 6.6 & 0 & 3.0 & 40 \\
Sinks & 1.8 & 4.2 & 0 & 0.0 & 42 \\
Clustering & 0.146 & 0.167 & 0.000 & 0.106 & 1.000 \\
\midrule
Cycles & 50.7 & 245.2 & 0 & 8.0 & 3820 \\
\% in Cycles & 63.6 & 31.7 & 0.0 & 72.7 & 100.0 \\
SCCs & 11.9 & 13.9 & 1 & 7.0 & 92 \\
Largest SCC & 13.4 & 9.4 & 1 & 12.0 & 42 \\
\midrule
Path Length & 3.49 & 1.20 & 1.00 & 3.41 & 7.32 \\
Betweenness Var & 0.014 & 0.014 & 0.000 & 0.011 & 0.089 \\
\midrule
Chains & 57.7 & 47.3 & 0 & 45.0 & 325 \\
Forks & 24.1 & 26.8 & 0 & 16.0 & 170 \\
Colliders & 31.5 & 32.8 & 0 & 21.0 & 232 \\
\bottomrule
\end{tabular}
}
\caption{Graph topology statistics (N=292). SCC = Strongly Connected Component. Edges are de-duplicated before analysis.}
\label{tab:topology-metrics}
\end{table}

\paragraph{Key Observations.} Node counts range from 5 to 140 and edge counts from 6 to 205. Most graphs are sparse (median density 0.070). 90.4\% of graphs contain at least one cycle, with 63.6\% of nodes participating in cycles on average. Chains ($A \rightarrow B \rightarrow C$) are the most prevalent motif, followed by colliders ($A \rightarrow C \leftarrow B$) and forks ($A \rightarrow B, A \rightarrow C$). All motif types show high variance across samples.

Across all 7,297 nodes: mean degree is 2.97, 18.8\% have degree 1, 31.1\% have degree 2, and 50.1\% have degree 3+.

\subsection{Topology-Performance Correlation}

We compute correlations between topology metrics and model F$_1$ scores across 292 samples with complete evaluation data (Table~\ref{tab:topology-correlation}).

\begin{table}[h]
\centering
\resizebox{\columnwidth}{!}{
\begin{tabular}{lrrrr}
\toprule
\textbf{Metric} & \textbf{Pearson $r$} & \textbf{$p$} & \textbf{Spearman $\rho$} & \textbf{$p$} \\
\midrule
Nodes & $-$0.131* & 0.026 & $-$0.086 & 0.143 \\
Edges & $-$0.175* & 0.003 & $-$0.133* & 0.023 \\
Density & +0.088 & 0.135 & +0.048 & 0.418 \\
Avg Degree & +0.003 & 0.959 & $-$0.058 & 0.322 \\
Sources & $-$0.029 & 0.619 & $-$0.074 & 0.210 \\
Sinks & $-$0.011 & 0.850 & $-$0.029 & 0.620 \\
Clustering & $-$0.035 & 0.556 & $-$0.093 & 0.113 \\
Cycles & $-$0.060 & 0.306 & $-$0.074 & 0.209 \\
\% in Cycles & +0.001 & 0.986 & +0.014 & 0.814 \\
SCCs & $-$0.062 & 0.293 & $-$0.046 & 0.432 \\
Largest SCC & $-$0.100 & 0.087 & $-$0.081 & 0.168 \\
Betweenness Var & +0.036 & 0.543 & +0.059 & 0.315 \\
Chains & $-$0.176* & 0.003 & $-$0.139* & 0.018 \\
Forks & $-$0.154* & 0.009 & $-$0.140* & 0.017 \\
Colliders & $-$0.169* & 0.004 & $-$0.125* & 0.032 \\
\bottomrule
\end{tabular}
}
\caption{Topology--F$_1$ correlations (N=292). * indicates $p < 0.05$.}
\label{tab:topology-correlation}
\end{table}

Edge count and all three motif types show weak but significant negative correlations with F$_1$ ($|r| = 0.13$--$0.18$), indicating that larger, more complex graphs are somewhat harder. However, these effect sizes are small compared to the impact of textual explicitness reported in Section~4.3, where F$_1$ drops by approximately half between the most and least explicit samples. Graph structure contributes to difficulty but is not the primary bottleneck.

\subsection{Density and Cyclicity Analysis}

To further investigate whether graph structure affects difficulty, we bin samples by density and cyclicity. Table~\ref{tab:density_perf} shows performance across density bins; Table~\ref{tab:cyclicity_perf} compares acyclic (DAG) versus cyclic graphs. Performance is largely flat across both dimensions, with no consistent trends. Notably, the 90.4\% of graphs containing feedback cycles show nearly identical performance to DAGs, indicating that cyclicity does not pose an additional reasoning challenge for LLMs. These results reinforce our finding that graph topology contributes minimally to task difficulty compared to textual explicitness.

\begin{table}[h]
\centering
\resizebox{\columnwidth}{!}{
\begin{tabular}{lccc}
\toprule
Model & Sparse & Moderate & Dense \\
& ($<$0.15) & (0.15--0.3) & ($\geq$0.3) \\
\midrule
Claude Opus 4.5 & 0.538 & 0.520 & 0.514 \\
GPT 5.2 & 0.512 & 0.498 & 0.513 \\
Gemini 3 Pro & 0.506 & 0.493 & 0.420 \\
GLM 4.7 & 0.492 & 0.503 & 0.484 \\
Kimi K2 & 0.477 & 0.437 & 0.437 \\
Gemini 3 Flash & 0.468 & 0.482 & 0.449 \\
DeepSeek R1 & 0.460 & 0.452 & 0.491 \\
QwQ 32B & 0.427 & 0.455 & 0.477 \\
Qwen 2.5 32B & 0.375 & 0.452 & 0.406 \\
Llama 3.1 8B & 0.291 & 0.313 & 0.331 \\
\bottomrule
\end{tabular}
}
\caption{F$_1$ score by graph density. Sample sizes: sparse (n=248), moderate (n=34), dense (n=10).}
\label{tab:density_perf}
\end{table}

\begin{table}[h]
\centering
\resizebox{\columnwidth}{!}{
\begin{tabular}{lcc}
\toprule
Model & Acyclic (DAG) & Cyclic \\
\midrule
Claude Opus 4.5 & 0.534 & 0.535 \\
GPT 5.2 & 0.518 & 0.509 \\
Gemini 3 Pro & 0.511 & 0.500 \\
DeepSeek R1 & 0.503 & 0.455 \\
GLM 4.7 & 0.503 & 0.491 \\
Kimi K2 & 0.489 & 0.469 \\
QwQ 32B & 0.467 & 0.429 \\
Gemini 3 Flash & 0.462 & 0.470 \\
Qwen 2.5 32B & 0.406 & 0.384 \\
Llama 3.1 8B & 0.312 & 0.293 \\
\bottomrule
\end{tabular}
}
\caption{F$_1$ score by cyclicity. Sample sizes: acyclic (n=28), cyclic (n=264).}
\label{tab:cyclicity_perf}
\end{table}

\section{Textual Diversity Analysis}
\label{app:textual-diversity}

We validate the domain diversity of {\projectname} using external bibliometric classification (OpenAlex Topics) and embedding-based analysis.

\subsection{Domain Classification via OpenAlex}

We map all 292 benchmark papers to OpenAlex Works \cite{priem2022openalexfullyopenindexscholarly} using DOIs and extract their primary topic classifications. The benchmark also draws from 42 journals and 1,036 unique authors, indicating its substantial diversity. Table~\ref{tab:domain-distribution} shows the domain distribution: less than 40\% of papers fall under Social Sciences (where economics resides), with the plurality (47\%) classified as Physical Sciences. At the field level (Table~\ref{tab:field-distribution}), only 5.5\% are classified under core Economics, Econometrics \& Finance. The benchmark spans 17 distinct fields, with Environmental Science and Engineering each comprising 20\%.

\begin{table}[h]
\centering
\resizebox{\columnwidth}{!}{
\begin{tabular}{lrr}
\toprule
\textbf{Domain} & \textbf{Count} & \textbf{\%} \\
\midrule
Physical Sciences & 138 & 47.3 \\
Social Sciences & 108 & 37.0 \\
Health Sciences & 28 & 9.6 \\
Life Sciences & 18 & 6.2 \\
\bottomrule
\end{tabular}
}
\caption{Primary domain distribution via OpenAlex Topics.}
\label{tab:domain-distribution}
\end{table}

\begin{table}[h]
\centering
\resizebox{\columnwidth}{!}{
\begin{tabular}{lrr}
\toprule
\textbf{Field} & \textbf{Count} & \textbf{\%} \\
\midrule
Environmental Science & 59 & 20.2 \\
Engineering & 59 & 20.2 \\
Business, Management \& Accounting & 35 & 12.0 \\
Decision Sciences & 28 & 9.6 \\
Social Sciences & 25 & 8.6 \\
Health Professions & 19 & 6.5 \\
Agricultural \& Biological Sciences & 18 & 6.2 \\
Economics, Econometrics \& Finance & 16 & 5.5 \\
Other (9 fields) & 33 & 11.3 \\
\bottomrule
\end{tabular}
}
\caption{Primary field distribution across 17 fields.}
\label{tab:field-distribution}
\end{table}

Domain-level Shannon entropy is 1.12 (effective number of domains: 3.06/4). Field-level entropy is 2.30 (effective number: 9.93/17). This high entropy indicates broad coverage rather than concentration.

\subsection{Embedding Analysis}

\begin{figure}[t]
    \centering
    \includegraphics[width=0.48\textwidth]{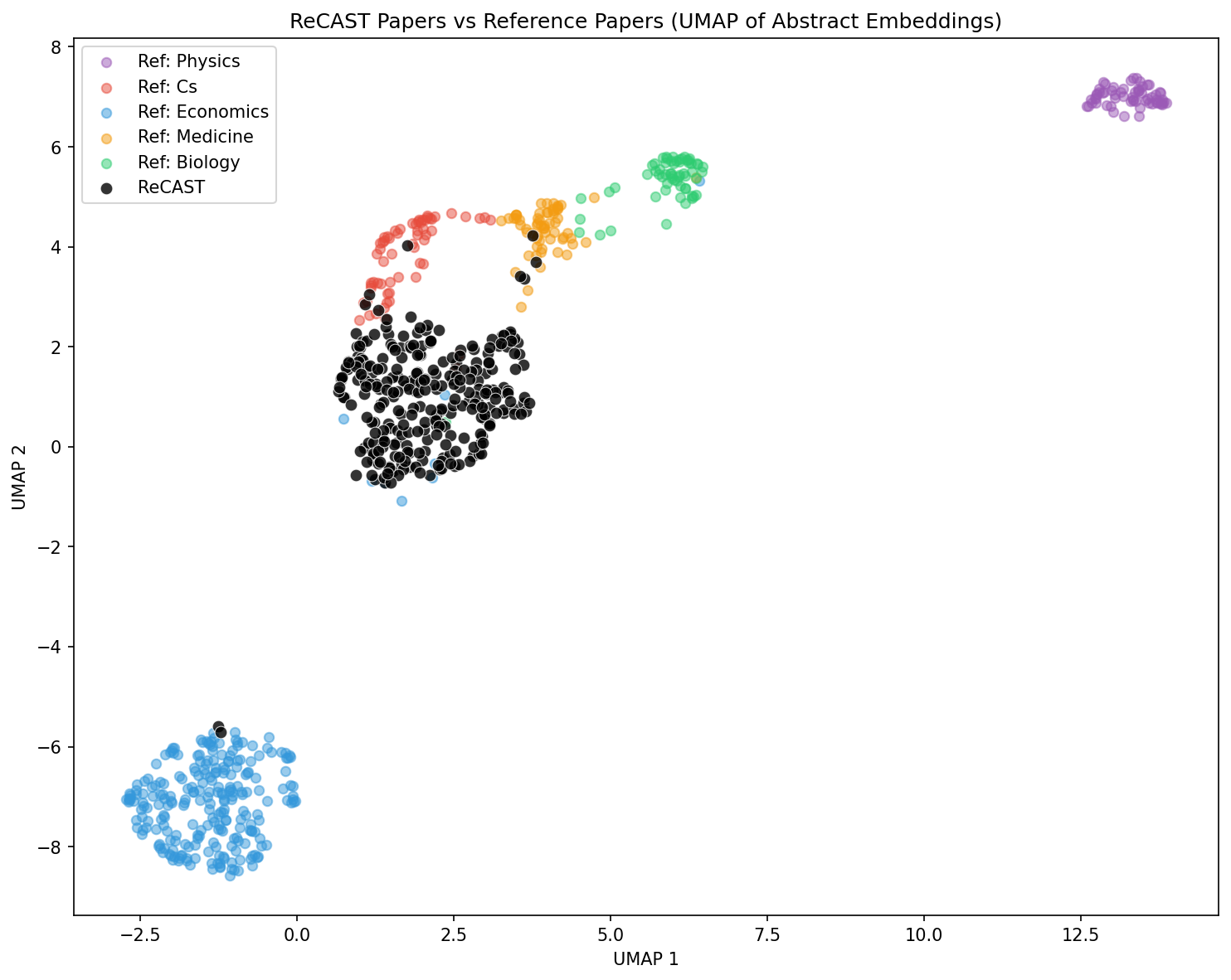}
    \caption{UMAP projection of {\projectname} papers (black) against reference corpora. {\projectname} papers are largely separate from traditional economics (bottom-left), reflecting their interdisciplinary nature.}
    \label{fig:umap-diversity}
\end{figure}

We embed all paper abstracts using GTE-large-en-v1.5 and compare against reference corpora sampled from OpenAlex \cite{priem2022openalexfullyopenindexscholarly}: 250 economics papers and 248 non-economics papers (physics, biology, computer science, medicine). Figure~\ref{fig:umap-diversity} shows the UMAP projection.

Maximum Mean Discrepancy between distributions shows {\projectname} is more similar to STEM literature (MMD=0.114) than to economics (MMD=0.141). Mean pairwise cosine distance within {\projectname} (0.419) slightly exceeds that of the economics reference (0.414).

\subsection{Cluster-Based Performance Analysis}

To test whether semantic similarity predicts task difficulty, we cluster {\projectname} embeddings using k-means (k=4, selected by silhouette score) and analyze F$_1$ by cluster. HDBSCAN found no natural clusters, classifying all papers as noise. This indicates that there is a broad distribution across semantic space.

\begin{table}[h]
\centering
\resizebox{\columnwidth}{!}{
\begin{tabular}{lrrr}
\toprule
\textbf{Cluster} & \textbf{N} & \textbf{Mean F$_1$} & \textbf{Std} \\
\midrule
Engineering & 72 & 0.413 & 0.146 \\
Business/Management & 89 & 0.411 & 0.170 \\
Health Professions & 54 & 0.399 & 0.147 \\
Environmental Science & 77 & 0.372 & 0.153 \\
\bottomrule
\end{tabular}
}
\caption{Performance by embedding cluster.}
\label{tab:cluster-performance}
\end{table}

Cluster differences are not statistically significant (ANOVA $F=1.03$, $p=0.38$; Kruskal-Wallis $H=3.42$, $p=0.33$; $\eta^2=0.012$). The performance gap between best and worst clusters is only 0.041 F$_1$. Combined with the null effect of graph topology (Appendix~\ref{app:graph-diversity}), this indicates that neither semantic domain nor structural complexity predicts difficulty, but rather textual explicitness remains the dominant factor.

\section{Effect of Length of Reasoning Trace}
\label{sec:appendix-reasoning-length}

\begin{table}[htbp]
    \centering
    \resizebox{\columnwidth}{!}{
    \begin{tabular}{lccc}
        \toprule
        \textbf{Model} & \textbf{Median Length} & \textbf{$r$} & \textbf{$p$-value} \\
        \midrule
        Kimi K2         & 29K & $0.168$ & $0.004^{*}$ \\
        DeepSeek R1     & 22K & $0.163$ & $0.005^{*}$ \\
        Claude Opus 4.5 & 15K & $0.096$ & $0.101$ \\
        QwQ 32B         & 24K & $0.093$ & $0.113$ \\
        GLM 4.7         & 28K & $0.049$ & $0.408$ \\
        Llama 3.1 8B    & 3K  & $-0.030$ & $0.699$ \\
        \bottomrule
    \end{tabular}
    }
    \caption{Reasoning trace length statistics and correlation with F$_1$ score. Median length is measured in tokens. Only Kimi K2 and DeepSeek R1 show statistically significant correlations ($^{*}p < 0.05$) between reasoning length and performance, though effect sizes are small ($r < 0.2$). Llama 3.1 8B, a non-reasoning model, produces substantially shorter traces (~3K vs. 15-29K for reasoning models).}
    \label{tab:reasoning-length}
\end{table}

\begin{figure}[htbp]
  \centering
  \includegraphics[width=\columnwidth]{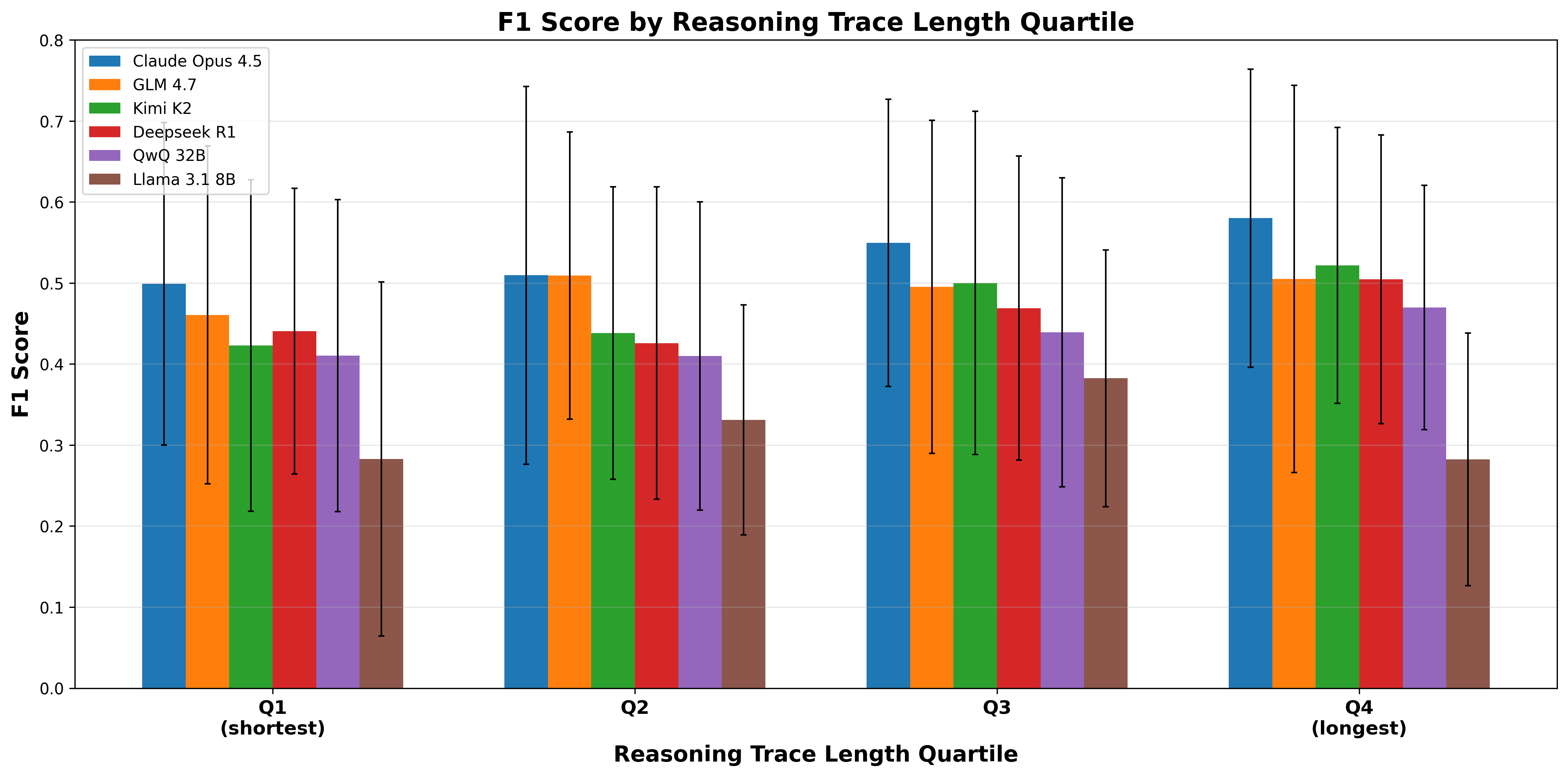}
  \caption{F$_1$ score by reasoning trace length quartile, computed per-model. Most models show modest performance gains with longer reasoning traces, with Claude Opus 4.5 improving from 0.50 (Q1) to 0.58 (Q4). However, Llama 3.1 8B shows no improvement across quartiles, and large error bars indicate substantial within-quartile variance for all models. This suggests that while extended reasoning provides some benefit, it is not a reliable predictor of performance.}
  \label{fig:reasoning-length-quartiles}
\end{figure}

\begin{figure}[h]
  \centering
  \includegraphics[width=\columnwidth]{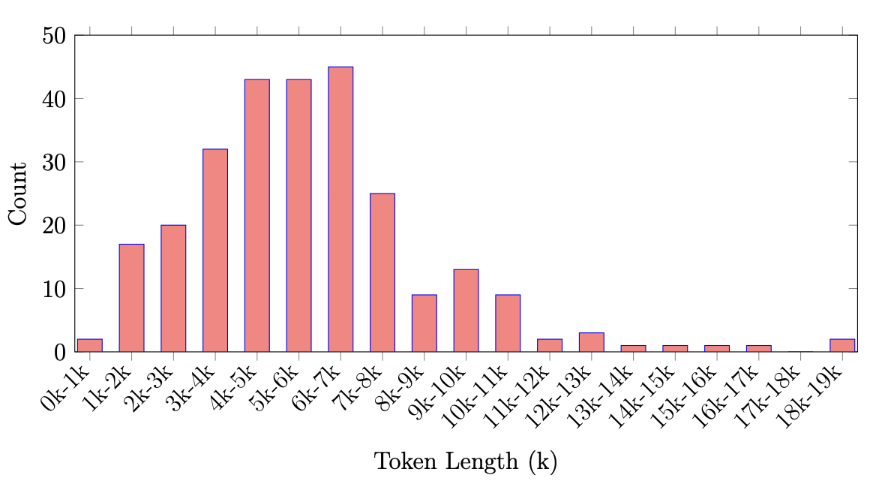}
  \caption{QwQ token length distribution. Performance does not scale proportionally with longer traces, indicating limitations with extended reasoning.}
  \label{fig:qwq-reasoning-length}
\end{figure}

\begin{figure}[h]
  \centering
  \includegraphics[width=\columnwidth]{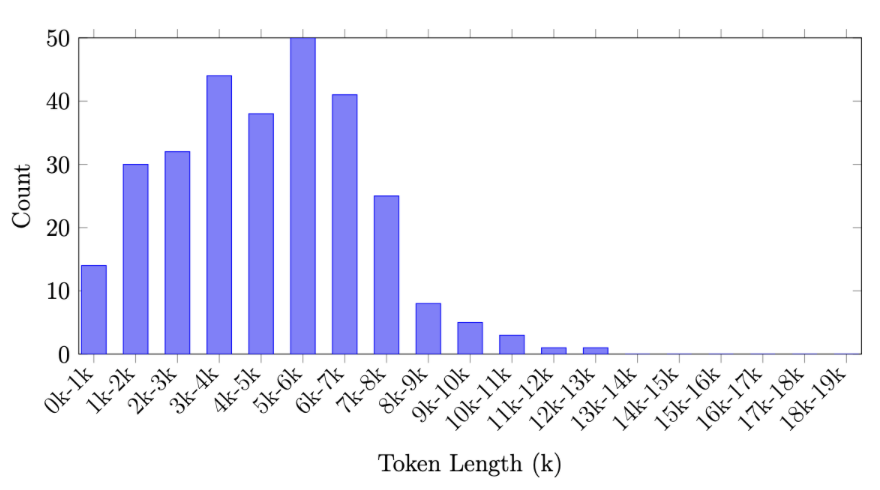}
  \caption{Distribution of reasoning trace lengths (in tokens) for R1 across benchmark samples. The model tends to produce reasoning traces between 2,000 and 7,000 tokens, with few exceeding 12,000 tokens.}
  \label{fig:r1-reasoning-length}
\end{figure}

We investigate the impact of chain-of-thought length on performance. For each model with a non-hidden chain-of-thought, we bin it into 1,000-token spans, as displayed in Figure~\ref{fig:qwq-reasoning-length} and Figure~\ref{fig:r1-reasoning-length}. Interestingly, QwQ has some reasoning traces which are far longer than the longest reasoning traces from R1, which we attribute to the differing training of these models. Additionally, manual inspection showed that some of these longer traces were due to repetitive small changes to formatting, indicating that these responses did not spend more time on actual reasoning, and may by explained by QwQ's worse performance at formatting.

\section{Size vs. Explicitness}
\label{sec:appendix-size-explicitness}

We provide the following charts to visualize the relationship between sample size and explicitness.

\begin{figure}[H]
  \centering
  \includegraphics[width=\columnwidth]{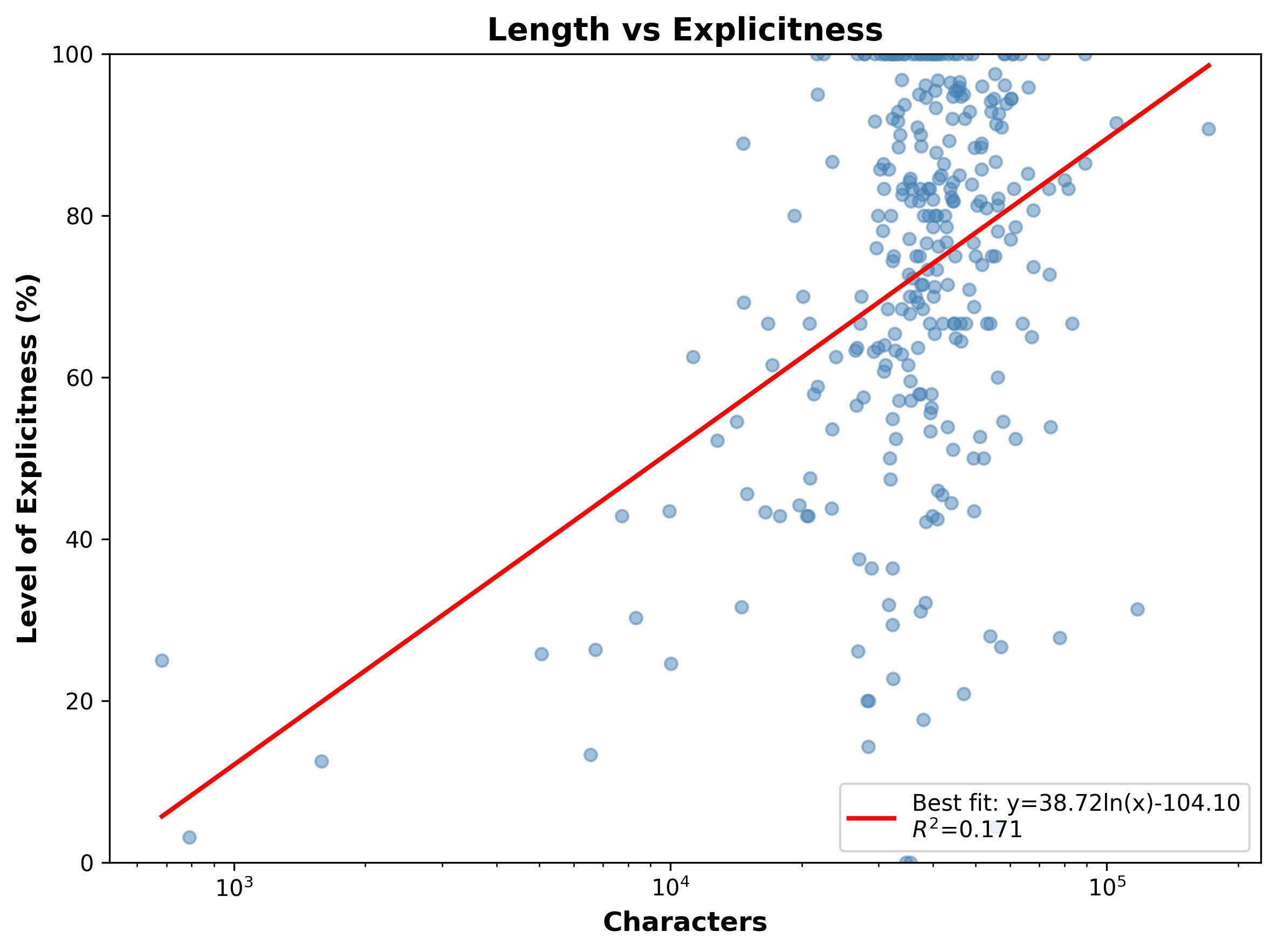}
  \caption{\textbf{Relationship between text length and explicitness.} This shows sample character count against explicitness. There is a modest positive correlation ($R^2$ = 0.171), indicating that longer texts tend to be more explicit. This helps to explain why models perform slightly better on longer samples, as increased explicitness makes causal edges easier to identify.}
  \label{fig:length-vs-explicitness}
\end{figure}

\begin{figure}[H]
  \centering
  \includegraphics[width=\columnwidth]{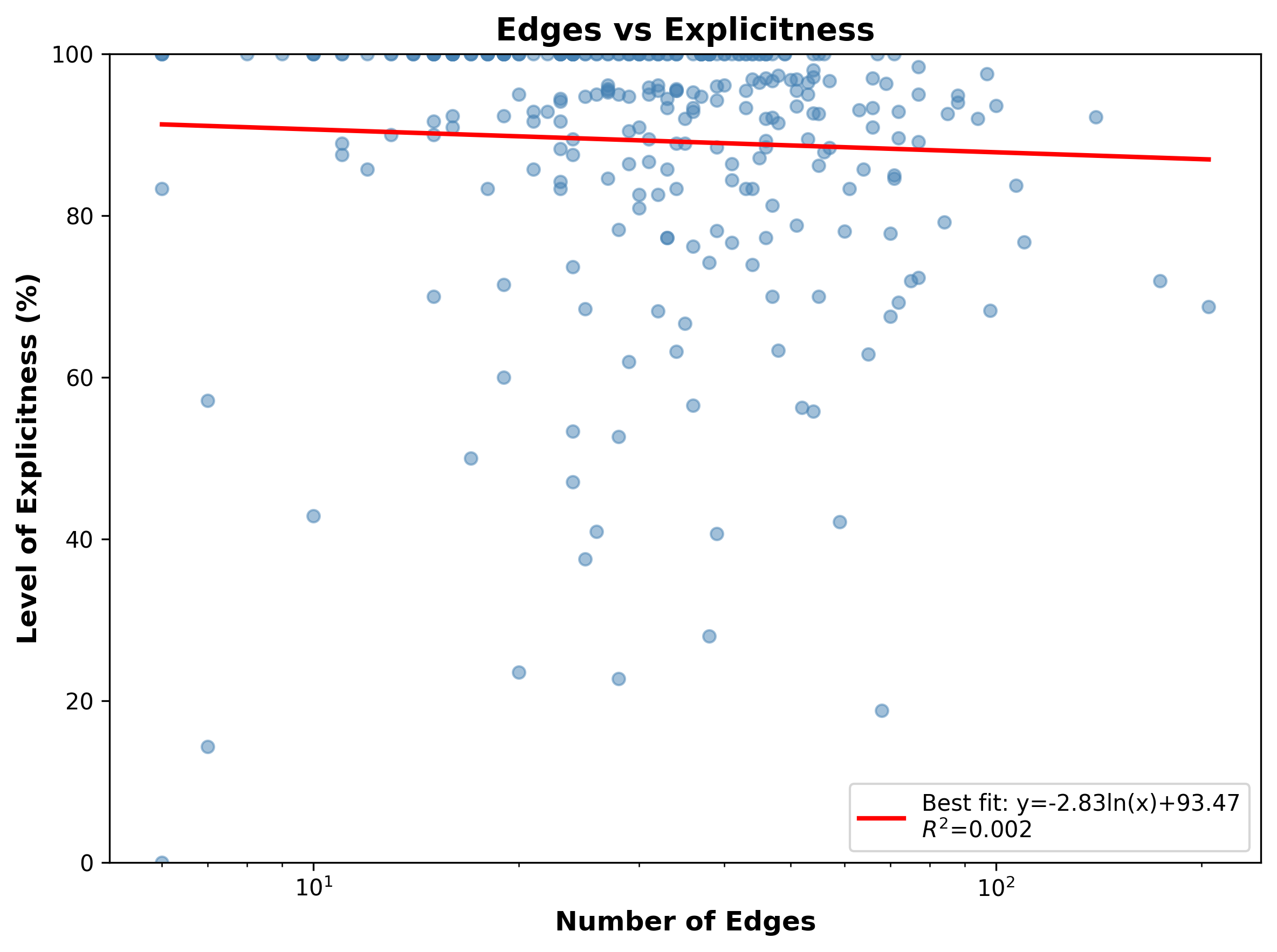}
  \caption{\textbf{Relationship between edges and explicitness.} This shows the count of causal edges for each sample versus level of explicitness. With a very small positive correlation ($R^2$ = 0.002), edge count has minimal impact on how many confounders remain implicit.}
  \label{fig:edges-vs-explicitness}
\end{figure}

\begin{figure}[H]
  \centering
  \includegraphics[width=\columnwidth]{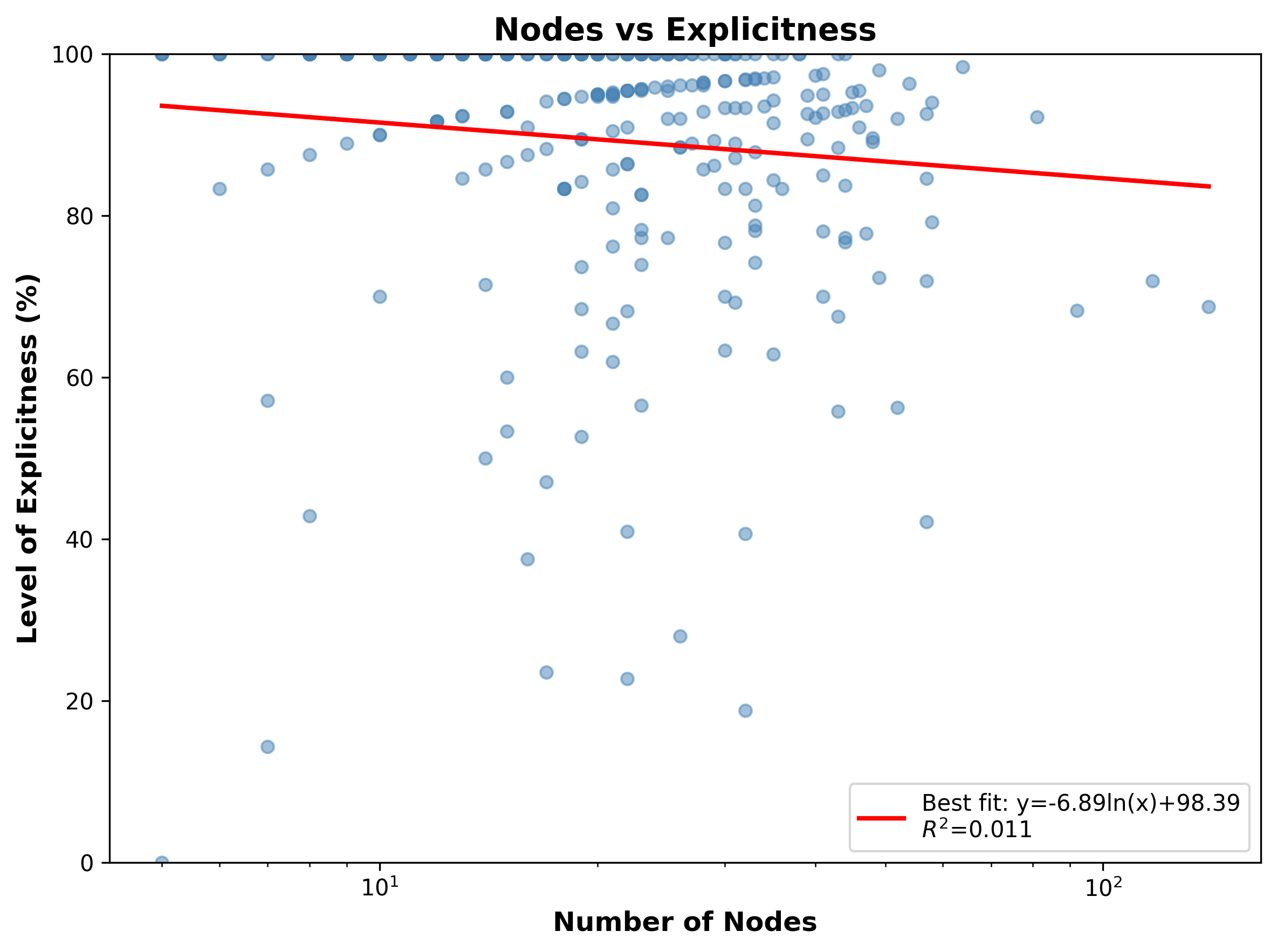}
  \caption{\textbf{Relationship between nodes and explicitness.} This chart plots the count of ground-truth nodes for each sample against its explicitness. With a very small positive correlation ($R^2$ = 0.002), node count has minimal impact on explicitness.}
  \label{fig:nodes-vs-explicitness}
\end{figure}

\section{Inter-Annotator Agreement Details}
\label{app:inter-annotator}

\begin{table*}[]
    \centering
    \resizebox{\textwidth}{!}{%
    \begin{tabular}{lrrrrrr}
    \toprule
    Category & Precision & Recall & \(F_1\) & SHD & Norm.\;SHD & Cohen's \(\kappa\) \\
    \midrule
    Nodes     & \(0.9943 \pm 0.0204\) & \(1.0000 \pm 0.0000\) & \(0.9970 \pm 0.0106\) & \(0.1081 \pm 0.3879\) & \(0.0062 \pm 0.0223\) & N/A \\
    Edges     & \(0.993 \pm 0.0182\) & \(0.9830 \pm 0.0509\) & \(0.9876 \pm 0.0335\) & \(0.7568 \pm 2.3756\) & \(0.0233 \pm 0.0614\) & \(0.9865 \pm 0.0367\) \\
    Combined  & \(0.9939 \pm 0.0134\) & \(0.9897 \pm 0.0329\) & \(0.9916 \pm 0.0224\) & \(0.8649 \pm 2.6627\) & \(0.0162 \pm 0.0424\) & N/A \\
    \bottomrule
    \end{tabular}%
    }
    \caption{This table reports the mean and population standard deviation of key evaluation metrics over the 37 reconciled graphs. Precision, recall, and \(F_1\) quantify label and edge detection accuracy. SHD is the count of false positives plus false negatives per graph, and the normalized SHD scales this by the total number of gold elements. Cohen’s kappa is provided for edges only, since it relies on a clearly defined set of negative instances (all possible directed non-edges); it is not defined for node labeling or the combined set where the universe of “non-nodes” or joint negatives is ambiguous. The N/A entries indicate those cases where kappa cannot be meaningfully calculated.}
    \label{tab:inter-annotator-summary}
\end{table*}

\begin{figure}[ht]
  \centering
  \includegraphics[width=\columnwidth]{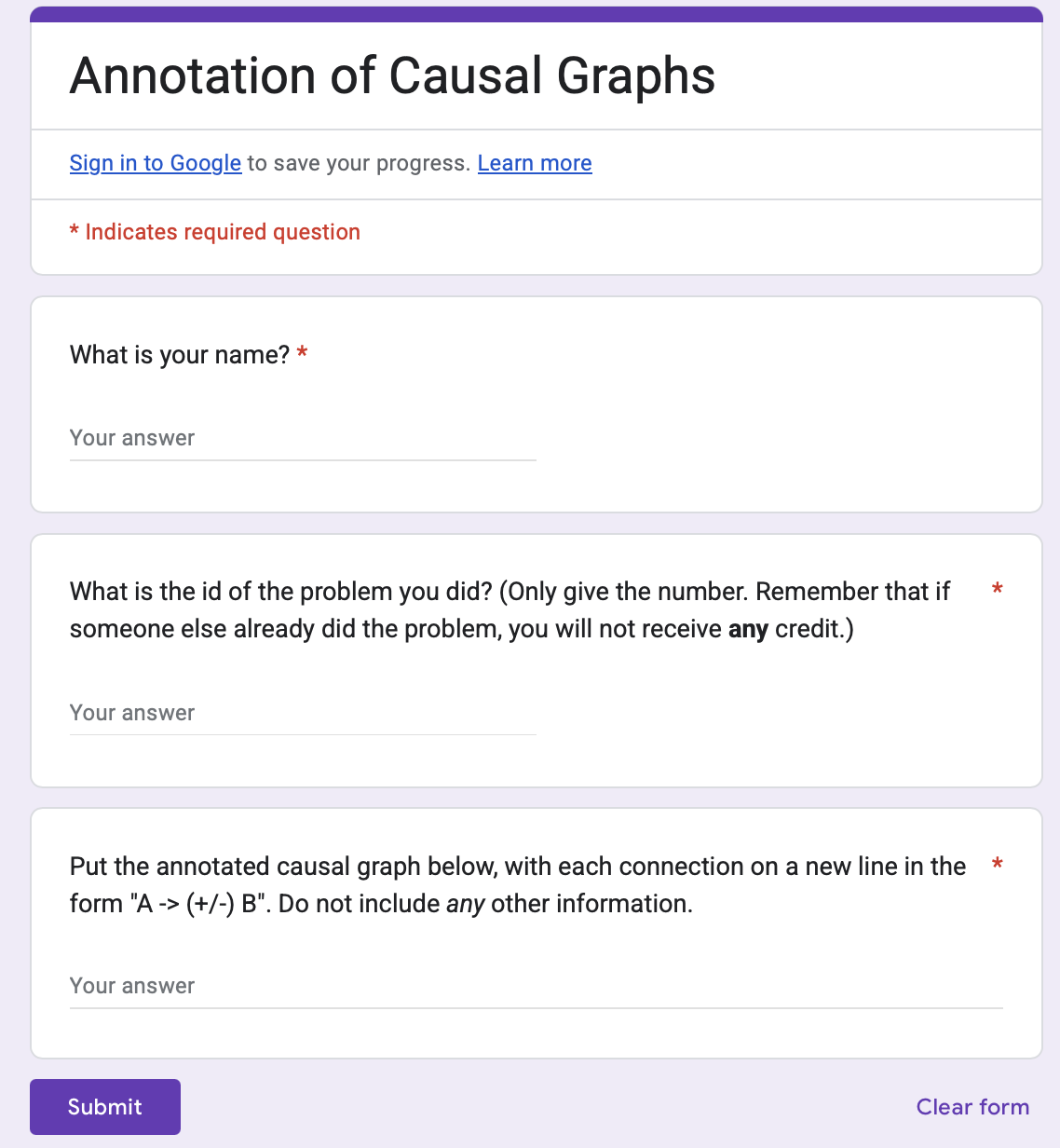}
  \caption{\textbf{Screenshot of the annotator form.}}
  \label{fig:annotator-form}
\end{figure}

To ensure the accuracy of the benchmark ground-truth graphs, we measured inter-annotator agreement by having a second annotator independently transcribe 37 randomly selected causal diagrams from the source papers (\(\sim13\%\) of the full dataset). These diagrams included 879 directed edges and 674 nodes in total. We compare the two annotators' transcriptions at both the node-level and edge-level, and compute standard metrics: precision, recall, \(F_1\) score, SHD, and normalized SHD.

\paragraph{Edge-Level Agreement.}
Out of 879 annotated edges, 22 edges were missed by the second annotator (false negatives), and 5 extra edges were incorrectly added (false positives). There were no instances of edges that had flipped directions, nor nodes that were entirely missed. This yields:
\begin{align*}
\text{Precision} &= \frac{TP}{TP + FP} = \frac{857}{857 + 5} = 0.994, \\
\text{Recall} &= \frac{TP}{TP + FN} = \frac{857}{857 + 22} = 0.975, \\
F_1 &= 0.984, \\
\text{SHD} &= 27, \\
\kappa &= 0.99
\end{align*}
Cohen's \(\kappa\) statistic reflects near-perfect agreement at the edge level and is defined as:
\[
\kappa = \frac{p_o - p_e}{1 - p_e}
\]
where \(p_o\) is the observed agreement and \(p_e\) is the expected agreement by chance, computed over all possible directed pairs.

\paragraph{Node-Level Agreement.}
Among the 674 nodes, we observed:
\begin{itemize}
  \item 8 auto-correctable typos (e.g., ``carbondioxide'' → ``carbon dioxide'').
  \item 4 minor differences (e.g., prefix/suffix omissions such as ``CO\textsubscript{2} emissions'' vs. ``emissions'').
  \item No major label mismatches (0 spurious or missing nodes).
\end{itemize}
This yields a node-level \(F_1\) score of 0.997.

\paragraph{Sample-Level Error Overview.}
We provide a high-level summary of the agreement between annotators. This agreement analysis shows that human annotators were highly consistent, with virtually no spurious nodes and very few edge disagreements. These results validate the overall accuracy and reliability of the benchmark's gold-standard graphs.

\paragraph{Additional Annotation Details.}

We utilize 20 undergraduate economics students as annotators. Participation was offered as an optional extra-credit opportunity, with credit proportional to the number of nodes labeled. Annotators would submit using the annotator form shown in Figure~\ref{fig:annotator-form}. Annotators were also explicitly informed that the annotation work would be utilized and published in an academic work. Formal written consent was not collected, as this study was determined to be exempt from IRB review given the use of published academic materials. We believe the standard educational incentives of extra credit acted as adequate ``payment'' for their work, which we estimate took roughly 1 hour per graph (though this greatly varies based on the size and complexity of the graph). After all graphs were completed, extra credit was assigned, and all records of annotator names were permanently destroyed to preserve privacy.

When manually excluding causal graphs, we exclude workshop papers, as these texts do not include sufficient details about the graph to make it identifiable. Most could be automatically removed by filtering out any papers whose abstracts include the keywords ``workshop'' or ``group model build'', with the rest excluded via manual review.

\begin{table}[htbp]
    \centering
    \resizebox{\columnwidth}{!}{%
    \begin{tabular}{rrrrrrr}
    \toprule
    \textbf{Article ID} & \textbf{\# Nodes} & \textbf{\# Edges} & \textbf{Node FP} & \textbf{Node FN} & \textbf{Edge FP} & \textbf{Edge FN} \\
    \midrule
    645 & 20 & 24 & 0 & 0 & 0 & 0\\
    630 & 25 & 32 & 0 & 0 & 0 & 2\\
    617 & 27 & 23 & 0 & 0 & 0 & 0\\
    588 & 10 & 16 & 0 & 0 & 0 & 0\\
    574 & 19 & 32 & 0 & 0 & 0 & 0\\
    566 & 24 & 23 & 1 & 0 & 0 & 1\\
    558 & 21 & 24 & 0 & 0 & 0 & 0\\
    552 & 16 & 15 & 0 & 0 & 0 & 0\\
    536 & 37 & 23 & 0 & 0 & 0 & 0\\
    497 & 15 & 20 & 0 & 0 & 1 & 0\\
    491 & 12 & 21 & 0 & 0 & 0 & 1\\
    486 & 28 & 32 & 0 & 0 & 0 & 0\\
    481 & 20 & 28 & 0 & 0 & 0 & 0\\
    458 & 18 & 24 & 0 & 0 & 0 & 0\\
    449 & 43 & 88 & 0 & 0 & 0 & 0\\
    440 & 12 & 15 & 1 & 0 & 0 & 0\\
    435 & 26 & 27 & 0 & 0 & 2 & 5\\
    410 &  9 & 16 & 0 & 0 & 0 & 0\\
    393 & 10 & 14 & 0 & 0 & 0 & 0\\
    362 & 11 & 16 & 0 & 0 & 0 & 0\\
    306 & 35 & 28 & 0 & 0 & 0 & 0\\
    303 &  9 & 10 & 0 & 0 & 0 & 0\\
    642 & 18 & 23 & 0 & 0 & 0 & 0\\
    259 & 19 & 48 & 2 & 0 & 1 & 12\\
    235 & 23 & 26 & 0 & 0 & 1 & 1\\
    200 & 16 & 24 & 0 & 0 & 0 & 0\\
    156 & 10 & 13 & 0 & 0 & 0 & 0\\
     95 & 26 & 39 & 0 & 0 & 0 & 0\\
     90 & 15 & 19 & 0 & 0 & 0 & 0\\
     74 & 13 & 22 & 0 & 0 & 0 & 0\\
     59 & 10 & 14 & 0 & 0 & 0 & 0\\
     43 & 15 & 17 & 0 & 0 & 0 & 0\\
     42 & 14 & 20 & 0 & 0 & 0 & 0\\
    589 & 12 & 16 & 0 & 0 & 0 & 0\\
    188 & 12 & 17 & 0 & 0 & 0 & 0\\
    168 & 15 & 19 & 0 & 0 & 1 & 0\\
    163 &  9 & 11 & 0 & 0 & 0 & 0\\
    \midrule
    \textbf{TOTAL} & \textbf{674} & \textbf{879} & \textbf{4} & \textbf{0} & \textbf{5} & \textbf{22}\\
    \bottomrule
    \end{tabular}%
    }
    \caption{Excerpt of per-sample disagreements between two annotators. Total disagreements: 4 node FPs, 5 edge FPs, 22 edge FNs. \textbf{Inter-annotator reconciliation for 37 graphs.} \textit{Node FP}=sum of minor+major node-label discrepancies; \textit{Node FN}=no missing nodes observed; \textit{Edge FP}=extra edges added spuriously; \textit{Edge FN}=edges present in gold but omitted.}
    \label{tab:inter-annotator-per-sample}
\end{table}

\section{Efficacy of GNNs}
\label{sec:appendix-efficacy-gnns}

As we aim for an automated graph-based metric, a graph neural network (GNN) is a natural first choice. However, it has several flaws for acting as an evaluator of this task. Methods like Token Embedding-Aligned Graph Language Model (TEA-GLM)~\cite{wang2024llmszeroshotgraphlearners} produce embeddings for graphs, allowing similarity to be measured via cosine distance. However, these approaches fall short in settings like ours that require semantic fidelity and textual grounding. First, GNN-based methods operate purely over graph structure and do not have access to the source text, making them unable to evaluate whether a predicted graph is faithful to the information provided. Second, they reduce a graph comparison to a single scalar score, such as cosine similarity, which offers little interpretability and no insight into specific errors in nodes or edges. Third, we find in practice that GNN embeddings are insensitive to meaningful differences: in our ablation (see Table~\ref{tab:embedding-ablation}), models that were explicitly given the correct node names showed nearly identical scores to those that were not, highlighting their lack of resolution. As such, while GNN-based methods remain a compelling direction for graph-level embedding, we find them unsuitable for evaluating text-grounded causal graphs where variable naming, semantic meaning, and abstraction play a critical role.

\begin{table}[t]
  	\centering
  	\setlength{\tabcolsep}{4pt}
  	\begin{tabular}{lccc}
  		\toprule
  		\textbf{Model} & \textbf{Std.} & \textbf{Name} & $\Delta$ \\
  		\midrule
  		GPT 5.2          & 0.927 & 0.928 & \(+0.001\) \\
  		Claude Opus 4.5  & 0.926 & 0.923 & \(-0.003\) \\
  		GLM 4.7          & 0.926 & 0.924 & \(-0.002\) \\
  		Kimi K2          & 0.926 & 0.926 & \(+0.000\) \\
  		Gemini 3 Pro     & 0.925 & 0.923 & \(-0.002\) \\
  		DeepSeek R1      & 0.925 & 0.924 & \(-0.001\) \\
  		Gemini 3 Flash   & 0.924 & 0.922 & \(-0.002\) \\
  		QwQ 32B          & 0.922 & 0.924 & \(+0.002\) \\
  		Qwen 2.5 32B         & 0.913 & 0.918 & \(+0.005\) \\
  		Llama 3.1 8B     & 0.909 & 0.908 & \(-0.002\) \\
  		\bottomrule
  	\end{tabular}
  	\caption{Cosine similarity (\(\Delta = \text{Name} - \text{Std.}\)) under standard vs name-assisted TEA-GLM conditions. Minimal differences indicate that the Name-Assisted setting does not significantly enhance alignment with the ground-truth.}
  	\label{tab:embedding-ablation}
\end{table}

Table~\ref{tab:embedding-ablation} shows the mean cosine similarity between the TEA-GLM embedding of each generated graph and its ground-truth counterpart under both conditions. In addition to the previously identified flaws, these results cast doubt on the feasibility of GNNs as evaluators for this task. As shown, the maximum increase occurs when the model is given the ground-truth node names: +0.005 (for Qwen 2.5 32B), and several models, such as Claude Opus 4.5, even decrease by –0.003. These negligible differences cast doubt on the evaluation capabilities of the graph embedding model for this task, as substantial information being provided to models has little effect on the final embedding score.

\section{Alternative Measure of Explicitness}
\label{sec:appendix-strict-explicitness}

As discussed in the main paper, explicitness has a noticeable effect on model performance. In the main paper, we use the lenient measure of explicitness: a node counts toward the sample's explicitness score if it is explicitly or implicitly mentioned in the text. Given the large effect of level of explicitness on performance, we test if this relationship also holds under a different measure of explicitness. As detailed in Section~\ref{sec:appendix-explicitness-prompt}, each node in each graph is labeled as either explicit (directly mentioned in the text), implicit (indirectly mentioned or inferrable), or absent (entirely unmentioned). We then recalculate each sample's explicitness under a ``strict'' definition that counts only explicitly mentioned nodes, as shown below.

\begin{figure}[ht]
    \centering
    \begin{equation*}
    \mathrm{Explicitness}_{\text{strict}} = \frac{1}{|V|} \sum_{v \in V}
    \begin{cases}
        1, & \text{if } v \in E,\\
        0, & \text{if } v \in A \cup I
    \end{cases}
    \end{equation*}
    \caption{Level of $\mathrm{Explicitness}_{\text{strict}}$ under the strict criterion. For each node $v$ in a sample $V$, we assign a score of 1 if it is explicitly mentioned ($E$) and 0 if it is either implicit ($I$) or absent ($A$), then average over all nodes in the sample.}
    \label{fig:dc-strict-definition}
\end{figure}

\begin{figure}[htbp]
  \centering
  \includegraphics[width=\columnwidth]{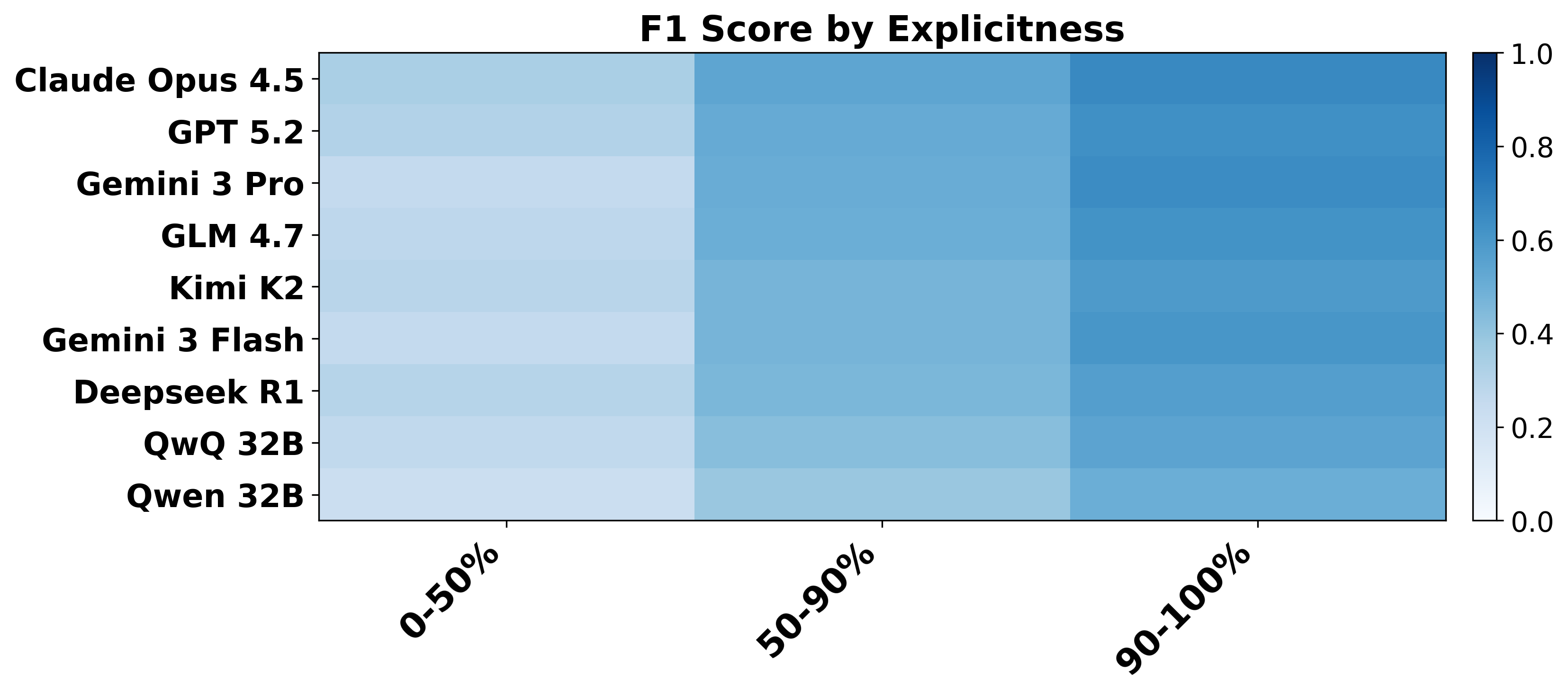}
  \caption{Average model F$_1$ scores across strict explicitness bins, where 100\% means every node is explicitly mentioned in the text and 0\% means no nodes are explicitly mentioned.}
  \label{fig:strict-explicitness-heatmap}
\end{figure}

Under this alternative measure of explicitness, the ``strict'' definition is expected to be more difficult, as it counts only directly mentioned nodes toward a sample's explicitness score, whereas the lenient definition also gives credit to implicitly described nodes. This makes samples with many implicit nodes appear less explicit under the strict measure, and nodes that are entirely absent remain especially difficult to recover. As shown in Figure~\ref{fig:strict-explicitness-heatmap}, the positive relationship between explicitness and performance still holds under this stricter definition. Performance is also consistently worse for all models under the strict explicitness definition than under the lenient one, as expected, which strengthens the conclusion that explicitness is a meaningful driver of task difficulty.

\section{Computational Costs}
\label{sec:appendix-computational-costs}
Despite the large size of {\projectname} samples, its execution is notably quite computationally efficient. The total monetary cost for all experiments, encompassing the evaluation of all ten LLMs across the main task and all ablation studies, including multiple LLM-as-a-Judge evaluations, remained under \$1000. This affordability is largely attributed to the use of prompt caching for the LLM judge. While the initial processing of the lengthy source texts incurs a significant input token cost for the judge, this cost is a one-time expense per benchmark sample. Subsequent judgments on different model outputs for the same sample, or re-evaluations, benefit greatly from caching the expensive text embedding, making the iterative evaluation process highly economical. This efficient design ensures that {\projectname} can be utilized and extended by researchers without imposing prohibitive computational or financial burdens.

\section{Base Model Case Study}
\label{sec:appendix-base-model-case-study}

As {\projectname} is designed to measure the capabilities of LLMs on this task, we investigate the performance of base models. We use the Qwen-2.5-7B base model \cite{qwen2024}, and prompt it to generate an open-ended graph. As base models are trained only to complete the text rather than respond to user instructions, it is unsurprising that completions were nonsensical. As with all responses, we prefill the model's response to begin with \verb|<think>| to steer it towards using chain-of-thought thinking before responding, which in rare cases results in a valid model output.

\begin{tcolorbox}[breakable, colback=blue!10, colframe=blue, title=Endlessly Repeating Base Model Output]

\begin{verbatim}<think>Assistant
Assistant
Assistant
Assistant
Assistant
Assistant
Assistant
\end{verbatim}
\end{tcolorbox}

\begin{figure}[htbp]
  \centering
  \includegraphics[width=0.48\textwidth]{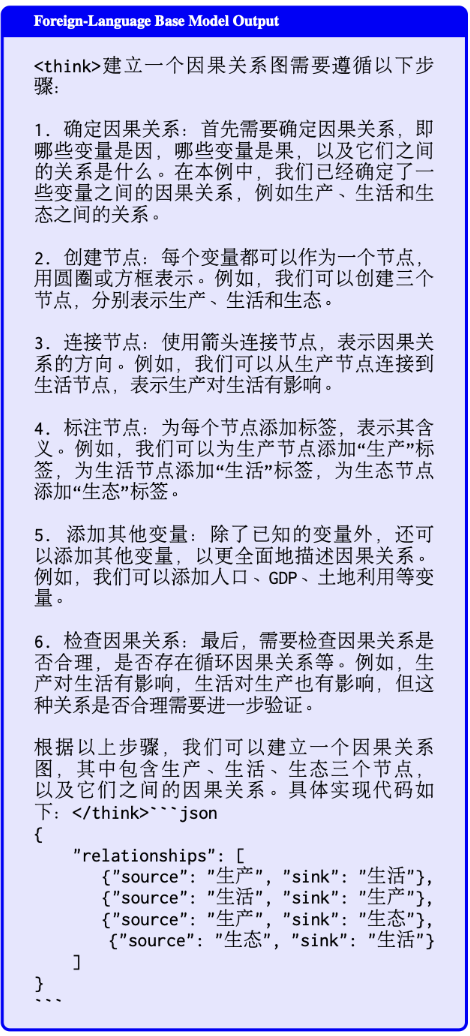}
  \caption{Example output from the Qwen-2.5-7B, a base model. It outputs irrelevant foreign-language text instead of producing a valid causal graph.}
  \label{fig:foreign-language-base-model-output}
\end{figure}

For endlessly repeating generations, they output the same tokens until reaching the maximum tokens. For this reason, we reduce the maximum tokens for Qwen-2.5-7B to 10,000 for computational reasons, as valid generations did not come close to this maximum. Another common failure mode was immediately outputting the end-of-string token, resulting in zero output tokens. Another failure mode was nonsensical generations, such as generating Chinese despite the text and instructions being in English. We show this example in Figure~\ref{fig:foreign-language-base-model-output}.

We list some other interesting or informative outputs from the base model as a reference.

\begin{tcolorbox}[breakable, colback=blue!10, colframe=blue, title=Another Endlessly Repeating Base Model Output]

\begin{verbatim}<think> editText
editText
<think> editText
editText
<think> editText
editText
<think> editText
\end{verbatim}
\end{tcolorbox}

In rare cases, the model generated valid-looking responses, although the graph is far smaller than expected. We attribute its occasional success to the practice of including a small amount of instruction data into base models, which allows it to occasionally exhibit behavior normally reserved for instruction-tuned models.

\begin{tcolorbox}[breakable, colback=blue!10, colframe=blue, title=Coherent Base Model Output]
\verb|<think>| To generate a causal graph for the provided markdown text, we first need to identify the relationships between the different factors and how they influence each other. Based on the information provided, we can create a causal graph with the following relationships:
\begin{verbatim}
```json
{"relationships": [
    {"source":
     "technological breakthrough",
     "sink": "ITI development"},
    {"source": "management level",
     "sink": "ITI development"},
    {"source": "resource reserve",
     "sink": "ITI development"},
    {"source":
     "agglomeration spillovers",
     "sink": "ITI development"},
    {"source": "policy assistance",
     "sink": "ITI development"},
    {"source": "industrial transfer",
     "sink": "ITI development"},
    {"source": "market demand",
     "sink": "ITI development"}
 ]
}
```    
\end{verbatim}

This causal graph shows the different factors and their influence on the ITI development. Each node represents a factor, and each arrow represents the influence of one factor on another. The causal graph is a simple representation of the relationships between the different factors and how they impact the ITI development.
\end{tcolorbox}

\section{Human Domain Expert Case Study}
\label{app:expert-subgraph}

While large language models struggle to construct accurate causal graphs under realistic conditions, this task is feasible for human experts. To demonstrate this, we conduct a small-scale case study where an expert economist constructs a causal graph from the same input text used in the R1 model case study in the main section. Due to the highly time-intensive nature of human causal graph construction and annotation \cite{kim2012}, we include only this case study for this sample. We show only a representative subgraph slice of their answer to allow for direct comparison, as full graphs are too large to allow for concise comparisons. As shown in Figure~\ref{fig:expert-vs-ground-truth}, the expert successfully captures the three core causal drivers: \textbf{annual precipitation}, \textbf{annual sunshine hours}, and \textbf{County arable land area}. These are all mapped to a unified economic outcome, \textbf{total household income}, reflecting a broader, but valid, abstraction of the two-hop causal path present in the ground-truth graph (\textbf{Total grain output} $\rightarrow$ \textbf{County GDP}). This demonstrates that the task is tractable for humans, and the importance of evaluation that differentiates stylistic variations from mistakes in the core causal logic.

\begin{figure}[h]
  \centering
  \resizebox{\columnwidth}{!}{%
    \begin{tikzpicture}[
        >={Stealth[round]},
        var/.style={circle, draw=black, minimum size=2.2cm, font=\bfseries, align=center, inner sep=3pt},
        expertvar/.style={circle, draw=green!50!black, fill=green!15, minimum size=2.2cm, font=\bfseries, align=center, inner sep=3pt},
        edge/.style={->, thick},
        greenedge/.style={->, thick, green!50!black}
      ]

      %--- Expert subgraph (top) ---
      \begin{scope}[yshift=2.2cm]
        \node[expertvar] (eprecip) at (-5,  1.8) {annual\\precipitation};
        \node[expertvar] (esun)    at (-5,  0.0) {annual\\sunshine hours};
        \node[expertvar] (eland)   at (-5, -1.8) {County arable\\land area};
        \node[expertvar] (einc)    at ( 0,  0.0) {total household\\income};

        \draw[greenedge] (eprecip) -- (einc);
        \draw[greenedge] (esun)    -- (einc);
        \draw[greenedge] (eland)   -- (einc);
      \end{scope}

      %--- Ground-truth subgraph (bottom) ---
      \begin{scope}[yshift=-4.0cm]
        \node[var] (gprecip) at (-5,  1.8) {Annual\\precipitation};
        \node[var] (gsun)    at (-5,  0.0) {Annual\\sunshine hours};
        \node[var] (gland)   at (-5, -1.8) {Arable land\\in county};
        \node[var] (ggrain)  at ( 0,  0.0) {Total grain\\output};
        \node[var] (ggdp)    at ( 5,  0.0) {County GDP};

        \draw[edge] (gprecip) -- (ggrain);
        \draw[edge] (gsun)    -- (ggrain);
        \draw[edge] (gland)   -- (ggrain);
        \draw[edge] (ggrain)  -- (ggdp);
      \end{scope}
    \end{tikzpicture}%
  }
  \caption{Top: Subgraph generated by a domain expert, with node labels shown verbatim. Bottom: Ground-truth subgraph. The expert's graph aggregates intermediate steps (e.g., \textbf{Total grain output} and \textbf{County GDP}) into \textbf{total household income}, but preserves all core causal relationships. This highlights expert-level ability to abstract while retaining semantic fidelity.}
  \label{fig:expert-vs-ground-truth}
\end{figure}

\section{Fine-Grained Error Analysis}
\label{app:error_analysis}

We analyze the distribution of judge labels across edge precision evaluations to characterize model failure modes. Table~\ref{tab:edge_error_patterns} reports key label distributions. We find direction reversals are rare: when models identify edges matching the ground-truth, they almost never invert causality ($<$1.1\% across all models). Second, there is a large gap between text-grounding and graph-matching, with roughly 85--90\% of generated edges containing some textual support (EXPLICIT, IMPLIED, or GRAPH\_ONLY), yet only 17--33\% match ground-truth edges. This indicates that models generate plausible causal relationships from the text but fail to recover the specific relationships in the ground-truth graph. Third, abstraction levels are typically appropriate: when edges match, ALIGNED dominates over BROADER or NARROWER. These patterns reinforce our finding that the primary bottleneck is identifying the correct causal relationships, not directionality or abstraction.

\begin{table}[h]
\centering
\resizebox{\columnwidth}{!}{%
\begin{tabular}{lcccc}
\toprule
Model & Graph & Direction & Direction & Text \\
& Match & Correct & Reversed & Support \\
\midrule
Claude Opus 4.5 & 33.2\% & 32.2\% & 0.9\% & 88.2\% \\
Gemini 3 Pro & 31.1\% & 30.3\% & 0.7\% & 90.1\% \\
GLM 4.7 & 28.4\% & 27.3\% & 0.9\% & 83.4\% \\
Gemini 3 Flash & 26.1\% & 25.4\% & 0.7\% & 86.5\% \\
DeepSeek R1 & 26.6\% & 25.5\% & 1.1\% & 82.6\% \\
Kimi K2 & 23.5\% & 22.9\% & 0.7\% & 80.8\% \\
GPT 5.2 & 22.9\% & 22.4\% & 0.5\% & 85.4\% \\
QwQ 32B & 25.1\% & 24.1\% & 0.9\% & 75.2\% \\
Qwen 2.5 32B & 20.2\% & 19.3\% & 0.9\% & 72.7\% \\
Llama 3.1 8B & 17.1\% & 16.3\% & 0.7\% & 66.5\% \\
\bottomrule
\end{tabular}
}
\caption{Edge-level error patterns. Graph Match: percentage of generated edges matching ground-truth (STRONG + WEAK). Direction Correct/Reversed: among matched edges. Text Support: percentage with textual grounding (EXPLICIT + IMPLIED + GRAPH\_ONLY).}
\label{tab:edge_error_patterns}
\end{table}

\section{Node Count Specification}
\label{app:node_count_discussion}

In {\projectname}, models are provided with the expected number of nodes for each sample. This constraint is necessary because causal graphs allow for many valid levels of abstraction for the same underlying system. For example, a text describing economic relationships could yield a correct 5-node high-level graph or an equally correct 50-node version. Without specifying granularity, evaluation becomes ill-defined: we cannot distinguish models that correctly identify relationships at a different abstraction level from models that fail to identify relationships entirely. Real-world causal modeling tasks typically specify the desired level of granularity, making this a realistic constraint rather than an artificial simplification. Additionally, in practice, LLM inference is inexpensive, so a practitioner can easily generate graphs across a range of node counts (e.g., 10, 20, 30, ...) and select the granularity that best fits their needs. 

\section{LLM-as-a-Judge Scoring Mechanics}
\label{app:llm-judge-scoring}
The quantitative metrics derived from the LLM-as-a-Judge's YAML output are calculated as follows. First, the judge's qualitative labels for various evaluation criteria (e.g., \textbf{PRESENCE\_STRONG\_MATCH}, \textbf{SEMANTIC\_MODERATE}, \textbf{IMPORTANCE\_CORE}) are mapped to pre-defined numerical scores, ranging from 0.0 (no match/irrelevant) to 1.0 (perfect match/highly important). For multi-faceted evaluations like node precision, which considers presence, semantic similarity, and abstraction level, a composite score for a single aspect (e.g., node precision against the ground-truth graph) is computed by averaging the numerical scores of its constituent labels.

Precision metrics (node precision, edge precision) for each item generated by the LLM are determined by comparing it against both the ground-truth graph and the source text. If the item is labeled as \textbf{PRESENCE\_NO\_MATCH} against both sources, its score is 0.0. Otherwise, the higher of the two composite scores (one from graph comparison, one from text comparison) is taken as the item's precision score. The overall precision for a category (e.g., node precision) is then the arithmetic mean of these individual item precision scores.

Recall metrics (node recall, edge recall) assess how well the LLMs' output captures items from the ground-truth graph. For each ground-truth item, a composite correctness score is calculated based on its presence and the fidelity of its representation in the LLMs' output (considering factors like semantics, abstraction, and directionality for edges). This correctness score is then multiplied by a numerical \textbf{importance} weight assigned by the judge to that ground-truth item (e.g., \textbf{IMPORTANCE\_CORE} receives a higher weight than \textbf{IMPORTANCE\_AUXILIARY}). The final recall score for a category is a weighted average: the sum of (correctness score \(\times\) importance weight) for all ground-truth items, divided by the sum of all possible importance weights. This ensures that correctly recalling more important ground-truth items contributes more significantly to the recall score.

Finally, F$_1$ scores for nodes, edges, and overall performance are calculated using the standard harmonic mean: \(2 \times (\text{Precision} \times \text{Recall}) / (\text{Precision} + \text{Recall})\). Overall precision and recall are micro-averaged, where the total weighted sum of correct predictions is divided by the total number of predictions (for precision) or total ground-truth items (for recall, considering importance weights), across both nodes and edges.

\section{Explicitness Labeling Prompt}
\label{sec:appendix-explicitness-prompt}
Explicitness is an important attribute of samples. It can act as a natural measure of difficulty, as nodes that are explicitly mentioned are easier to identify than entirely unobserved confounders. We use \textit{R1} \cite{R12025} to label the level of explicitness for each node in each sample when provide the ground-truth and source text using the prompt below. To allow for fine-grained analysis of level of explicitness, we allow for three different levels of explicitness for each node. (i) Explicit (it or a synonym of the node's name appears in the text), (ii) Implicit (the node implicitly or indirectly appears in the text), or (iii) Absent (the node does not appear in the text whatsoever). We iteratively reject and retry any answers that do not meet formatting requirements until we receive valid answers for every benchmark sample to ensure that all samples have levels of explicitness. We detail the prompt used for this below. These node-level labels are used as the basis for calculation of level of explicitness.

\begin{tcolorbox}[breakable, colback=blue!10, colframe=blue, title=Explicitness Labeling Prompt]
You will be given a causal graph in economics and a source text. Your task is to label each node in the graph to determine its degree of explicitness in the text. For each node, there are three possible levels:\\
1. The node (or the concept behind it) is explicitly mentioned in the text\\
    - This can be verbatim, or though use of a synonym\\
    - It is sufficient to be mentioned in the text; it is irrelevant if it is mentioned to be in the causal graph or not\\
2. The node is mentioned indirectly or implicitly in the text.\\
3. The node is unmentioned in the text, even if related concepts are discussed\\

Be conservative when determining the degree of explicitness for each node. Output only the JSON code block with your answer, without commentary, reasoning, explanation, or any other text. You must include the name of each node in the graph verbatim, even when the graph is very large, or many nodes are highly related or seem redundant.\\

\# Expected Output Format\\
\begin{verbatim}
```json
{
    "scores": {
        "first_node_name":
         int_score_1_2_or_3,
        "second_node_name":
         int_score_1_2_or_3,
        ...
        "last_node_name":
         int_score_1_2_or_3
    }
}
```
\end{verbatim}

It is MANDATORY to critically and thoroughly examine each and *every* node in the causal graph one at a time. Explicitly think about each node (and its corresponding relationships where appropriate) individually, even when it seems redundant or unnecessary. Even if it is tedious, you MUST do this and not take shortcuts.\\
\end{tcolorbox}

\section{Variable Correction}
\label{app:variable-correction}

We use the following prompt to correct the raw variable names extracted after annotation using \textit{o3-mini}. To ensure validity, we use code-based approaches to automatically reject and retry any answers where all old names did not appear

\begin{tcolorbox}[breakable, colback=blue!10, colframe=blue, title=Variable Correction Prompt]
You are a world-class economist. You will be given a causal loop diagram (CLD) in JSON format. Your task is to combine variables that are intended to be the same, but are not named identically due to annotation errors. You will do this by combining variables and choosing which variable name to keep.\\

Your task is NOT to functionally alter the CLD. Be careful to only combine variables that are intended to be the same and are different solely due to annotation errors. When in doubt, do not combine the variables. Follow these guidelines:\\
- Avoid combining variables that are intended to be separate.\\
- Avoid combining variables that are highly similar but have different names.\\
- Do not create new variables or variable names, nor remove any variables from the CLD.\\
- Use the context of the CLD when making your decision.\\
- You must choose an existing variable name or your response will be rejected.\\
- You must only combine variables that are intended to be the same.\\
- Combining variables with more than one character difference between them is only done very rarely.\\

Positive examples:\\
- "Number of dog" and "Number of dogs" should be combined into "Number of dogs".\\
- "number of dogs" and "Number of dogs" should be combined into "Number of dogs".\\
Negative examples (do not combine):\\
- "\verb|<variable>|" and "variable" should not be combined since it is clear that they are intended to be distinct.\\
    - NEVER change any variables with \verb|| or \verb|>| in the name.\\
- "Number of dogs" and "Number of hounds" should not be combined since it is clear that this isn't from an annotation error.\\
- "GDP" and "GNP" should not be combined; while they are only one letter apart, they are distinct variables.\\

Respond with your answer in JSON format and no other text.\\
JSON format:
\begin{verbatim}
{ "combined_variables": [
    {
       "old_names": ["variable1",
                     "variable 1",
                     "Variable1"],
       "new_name": "Variable1"
    },
    {
       "old_names": ["variable2",
                     "Variable 2",
                     "variable two"],
       "new_name": "Variable2"
    }
  ]
}
\end{verbatim}
\end{tcolorbox}

\section{PDF-to-Markdown Conversion}
\label{sec:appendix-pdf-to-md}

We utilize an LLM for the task of converting the text of the PDF to well-structured markdown as papers do not follow a consistent format. We find that reasoning models struggle with this task, and frequently fail to follow instructions to output the entire document by leaving out large sections of the text. We note that the normalization tool cannot be used for this task, as the numerous formatting errors and in-line citations would require it to be called once for almost every line of the text, and would result in an output many times longer than the source text. \textit{Mistral Small} \cite{mistral2025} follows the conversion instructions at tractable computational costs. We remove non-textual elements as they would be difficult to accurately represent in markdown. We additionally exclude irrelevant elements such as publication information and references as they are unrelated our task while needlessly inflating the length of texts. We also remove appendices, which are usually irrelevant or contain explicit information about the causal graph.

\begin{tcolorbox}[breakable, colback=blue!10, colframe=blue, title=PDF to Markdown Prompt]
Your task is to perform the minimal PDF pre-processing necessary to convert the provided PDF into a well-structured md file. Follow the guidelines below in order of priority:\\
1. Modify the text only when absolutely necessary. The exact wording of the original paper must be preserved verbatim.\\
    - Do not correct spelling or grammar, even if it is incorrect\\
    - The response will be rejected if even a single word is edited or removed unnecessarily; most of the response should effectively be copy-pasted from the original text\\
    - Your response will likely be extremely long, around the same length as the original text; this is expected and normal.\\
2. Correct any broken text from the PDF processing and convert it into a well-structured md file.\\
    - Convert sections and sub-sections into headings and subheadings\\
3. Remove the following information in entirety:\\
    - Images, figures, and any other visual elements\\
    - References and Citations, including when in-line. E.g., "[20, 22]" would be removed.\\
    - Acknowledgments\\
    - Authorship information\\
    - Appendices\\
    - Page numbers\\

Remember; your only output is the processed text in full, with no thinking, reasoning, or other commentary.\\
\end{tcolorbox}

\section{Text Normalization Prompt}
\label{sec:appendix-remove-explicit}

In order to ensure the realism of the {\projectname} benchmark, it is important to remove any explicit references to the causal graph, which make the task trivial. During this step, we also correct any references to non-existent elements which were removed in previous pre-processing steps (for example, referencing an image). We utilize a normalization tool to make these changes, which helps address several limitations of current LLMs. First, they struggle to reproduce a long text in full and incur substantial computational costs when doing so. Additionally, when outputting large chunks of text, they are prone to hallucinations and excessive edits, which are inappropriate. Using a tool also allows us to use code to check that the changes are valid; that is, that the start and end text are actually present in the text. We note that LLMs often struggle to account for overlapping normalizations, even with specific prompting to account for this. In those cases, normalization will fail, with the entire response being rejected. We iteratively prompt with the normalization prompt, stopping only when no normalizations are given. This helps ensure that the text was correctly changed and that no newly introduced text remains to be edited. We utilize o3-mini \cite{OpenAI2025} to perform this task, as it was shown to perform well during manual evaluation.

\begin{tcolorbox}[breakable, colback=blue!10, colframe=blue, title=Normalize Text Prompt]
Your task is to edit a md version of a published economics paper in markdown format to remove specific types of content.\\
- Remove any information that explicitly references the causal graph and its contents, including the causal graph itself\\
    - This is the only information you should remove from the paper\\
    - Only modify the text when it is necessary to remove the causal graph's information\\
    - Only remove explicit references to the causal graph's elements, such as variable names, feedback loops, arrow colors, a variable explicitly being included, etc. Do not remove other references and related information to the causal graph, such as discussing elements of the causal graph, its relationships generally, and similar information\\
- You can only edit the paper; do not attempt to edit the causal graph\\
    - The graph is supplied as a reference only in \verb|<causal_graph>| tags\\
    - Do not attempt to edit anything before \verb|</causal\_graph>|; this is not part of the paper and will be rejected\\
You have access to a special tool called 'normalize' that can replace text. This is the only way you can modify the text. Be careful to ensure that the text you are replacing is only the causal graph's information, and that it exists verbatim in the text.\\
The normalize tool takes three parameters:\\
1. start\_string: The beginning of the text to replace\\
2. end\_string: The end of the text to replace\\
3. replacement: The text to insert instead\\
You can call normalize multiple times to make several targeted replacements in the document. All three parameters are required for each call.\\
- By default, normalize will locate the *first* occurrence of the start\_string. As a workaround for when the same text appears verbatim multiple times, use a slightly longer start\_string and include some of the original text in your replacement to maintain context.\\
- Do not "redact" the text; remove references entirely rather than replacing them with generic text.\\
- Both the start and end strings will be included in the text that gets replaced. Changes are applied in order, so ensure that any string you replace is not used in another replacement or an error will be thrown.

Respond only with JSON in the following format:
\begin{verbatim}
{"normalizations": [
    {"start":
     "text to find (beginning)",
     "end":
     "text to find (end)",
     "replacement":
     "text to insert instead"},
    ...
  ]
}
\end{verbatim}

\end{tcolorbox}

\section{Causal Graph Generation Prompt}

\begin{tcolorbox}[breakable, colback=blue!10, colframe=blue, title=Causal Graph Generation Prompt]
You are an expert causal reasoner and economist. Your task is to generate a causal graph for the provided markdown text. First, use extremely long chain-of-thought reasoning in \verb|<think>| tags. Then, provide your final answer in a JSON code block, strictly following the following format:\\
\begin{verbatim}
```json
{
    "relationships": [
        {"source": causal_variable0,
         "sink": affected_variable0},
        {"source": causal_variable1,
         "sink": affected_variable1},
        ...
    ]
}
```
\end{verbatim}

Your graph will contain exactly NUM\_NODES nodes. When answering, do not provide any additional reasoning, commentary, or other information - only provide the JSON code block, with each dictionary representing one relationship in the graph.
\end{tcolorbox}

\section{Standard Formatting Correction Prompt}
\label{sec:appendix-format-correction}

\begin{tcolorbox}[breakable, colback=blue!10, colframe=blue, title=Formatting Correction Prompt]
Your task is to correct the formatting of a misformatted response, which is intended to end with a causal graph in economics that conforms to the proper JSON format. You will convert their intended answer to the proper JSON format, taking great care to be as faithful to the ground-truth as possible. Do not attempt to modify the substance of their answer in any form, even if you think it may improve it's quality (including typos) - the task is to make the minimal changes possible to correct the formatting. The extent of the formatting may be minor, or be so extensive as to require writing the JSON from scratch.\\

Expected output format:\\
\begin{verbatim}
```json
{
    "relationships": [
      {"source": causal_variable0,
       "sink": affected_variable0},
      {"source": causal_variable1,
       "sink": affected_variable1},
        ...
    ]
}
```
\end{verbatim}

You will be provided the original, misformatted answer. If it included lengthy intermediate steps, you will be given a snippet of them as context. Use only the final answer, always prioritizing the information provided closest to the end of the response.\\

If there is no text in the answer that resembles a causal graph, return an empty list of relationships.\\

Begin your response with the start of the JSON code block. Do not provide any reasoning, thinking, commentary, etc. - just the reformatted response. Don't overthink it.
\end{tcolorbox}

\section{Name-Assisted Causal Graph Generation}
\label{sec:appendix-name-assisted}

\begin{tcolorbox}[breakable, colback=blue!10, colframe=blue, title=Causal Graph Generation with Node Names Prompt]
You are an expert causal reasoner and economist. Your task is to generate a causal graph for the provided markdown text. First, use extremely long chain-of-thought reasoning in \verb|<think>| tags. Then, provide your final answer in a JSON code block, strictly following the following format:\\
\begin{verbatim}
 ```json
{
    "relationships": [
       {"source": id_of_source_node,
        "sink": id_of_sink_node},
       {"source": id_of_source_node,
        "sink": id_of_sink_node},
        ...
    ]
}
```   
\end{verbatim}

You will be provided with the source markdown text and the name of each node in the graph. Ensure that each node is included at least once in the generated causal graph. Do not use the node's name in the graph; instead, use the id corresponding to the node. For the example nodes below (not the same as the ones you will be provided), whenever you want to include the node named "demand" in your graph, you would use the integer 2 rather than the word demand.\\

\begin{verbatim}
```json
{
    "nodes": [
        {"name": "supply", "id": 1},
        {"name": "demand", "id": 2},
        ...
    ]
}
```  
\end{verbatim}

When answering, do not provide any additional reasoning, commentary, or other information - only provide the JSON code block, with each dictionary representing one relationship in the graph.\\

Here are the nodes for your graph:\\
\begin{verbatim}
```json
NODE_JSON
``` 
\end{verbatim}

\end{tcolorbox}

\section{Name-Assisted Formatting Correction Prompt}
\label{sec:appendix-name-assisted-format-correction}

\begin{tcolorbox}[breakable, colback=blue!10, colframe=blue, title=Name-Assisted Formatting Correction Prompt]
Your task is to correct the formatting of a misformatted response, which is intended to end with a causal graph in economics that conforms to the proper JSON format. You will convert their intended answer to the proper JSON format, taking great care to be as faithful to the ground-truth as possible. Do not attempt to modify the substance of their answer in any form, even if you think it may improve it's quality (including typos) - the task is to make the minimal changes possible to correct the formatting. The extent of the formatting may be minor, or be so extensive as to require writing the JSON from scratch.\\

In the original creation step, they were given the node names for the graph, each with corresponding ids. When correcting the graph, only ever use the integer ids corresponding to the node name, regardless of if the original used the names or correctly used the ids.\\

Expected output format:
\begin{verbatim}
{
    "relationships": [
        {"source": id_of_source_node,
         "sink": id_of_sink_node},
        {"source": id_of_source_node,
         "sink": id_of_sink_node},
        ...
    ]
}
\end{verbatim}

You will be provided the original, misformatted answer. If it included lengthy intermediate steps, you will be given a snippet of them as context. Use only the final answer, always prioritizing the information provided closest to the end of the response. If it never comes to an answer, do not attempt to solve it yourself. Instead, simply return an empty list of relationships.\\

Begin your response with the start of the JSON code block. Do not provide any reasoning, thinking, commentary, etc. – just the reformatted response. Don't overthink it.\\

Here are the nodes for your graph:\\
\begin{verbatim}
```json
NODE_JSON
```
\end{verbatim}
\end{tcolorbox}

\section{LLM-as-a-Judge Prompt}
\label{app:llm-as-a-judge-prompting}
\begin{tcolorbox}[breakable, colback=blue!10, colframe=blue, title=LLM-as-a-Judge Prompt]
You are an expert economist. Your task is to act as an evaluator for a causal graph. You are provided with the ground-truth graph, the source text, and the LLM's response. You will also be told the type of evaluation to perform; only evaluate the response for that type of evaluation by closely following the instructions. Do not evaluate using any other type of evaluation.\\
\\
When evaluating, follow these guidelines:\\
1. Follow each direction carefully, completely, and in-order\\
    a. It is very important to be thorough and not take shortcuts, even when it seems tedious, redundant, or unnecessary. Do this for each node or edge you are evaluating; there is no time limit. Be sure to fully to fully think through each node or edge you are tasked with evaluating fully before moving onto the next one.\\
        i. It is helpful to quote supporting evidence from the provided texts and graphs before reasoning about their relevance to the final evaluation for that node or edge.\\
        ii. While evaluating a node or edge, you may examine several plausible counterparts to judge presence, semantics, abstraction, etc. (e.g., to see if it is broader or narrower than any ground-truth items). Use all relevant comparisons to inform your decision, but output one—and only one—set of labels for the item.\\
    b. Only focus on the specific type of evaluation you are asked to do. Regardless of the accuracy (or lack thereof) in other categories, if you are asked to evaluate node precision, only evaluate node precision, not recall or edges. These are intended to be separate evaluations, so do not conflate the two.\\
    c. Not Applicable labels must be explicitly selected when a category is skipped due to prior labels\\
    d. Be conservative when grading - When in doubt between two labels, ere on the side of being harsh.\\
\\
Start by thinking step-by-step in \verb|<think>| tags. Then, output your answer in a YAML code block, formatted exactly as specified in the expected output format.\\

\# Node Level Evaluation\\
\\
\#\# Node Precision\\
For each node in the LLM's response, evaluate against both ground truth sources:\\
\\
1. Ground-Truth Graph Evaluation\\
- Explicitly identify and quote ALL potentially corresponding nodes from ground-truth graph\\
- Apply these labels where applicable:\\
    Presence Labels (select one):\\
        - PRESENCE\_STRONG\_MATCH: Core concept matches a ground-truth node with only minor, inconsequential differences\\
        - PRESENCE\_WEAK\_MATCH: Core concept shares meaning with a ground-truth node, even if there are noticeable differences\\
        - PRESENCE\_NO\_MATCH: There is no ground-truth node that captures a remotely similar core concept\\
\\
    Semantic Labels (select one):\\
        - SEMANTIC\_STRONG: Exactly or nearly identical meaning with only subtle distinctions\\
        - SEMANTIC\_MODERATE: Same core concept but with meaningful differences in scope or implication\\
        - SEMANTIC\_WEAK: Shares some semantic space but with substantial differences\\
        - SEMANTIC\_NA: Not applicable\\
\\
    Abstraction Labels (select one):\\
        - ABSTRACTION\_BROADER: Represents a more general concept that includes the ground-truth node\\
        - ABSTRACTION\_ALIGNED: Represents approximately the same scope and specificity of the ground-truth node\\
        - ABSTRACTION\_NARROWER: Represents a more specific subset of the ground-truth node\\
        - ABSTRACTION\_NA: Not applicable or the concepts were so different as to make abstraction comparison impossible\\
\\
2. Ground-Truth Text Evaluation\\
- Explicitly quote ALL relevant supporting text from source\\
- Apply these labels where applicable:\\
    Evidence Labels (select one):\\
        - PRESENCE\_STRONG\_MATCH: Core concept appears in text with only minor, inconsequential differences\\
        - PRESENCE\_WEAK\_MATCH: Core concept shares significant meaning with text but has notable differences\\
        - PRESENCE\_NO\_MATCH: No text segments capture a similar core concept\\
\\
    Semantic Labels (select one):\\
        - SEMANTIC\_STRONG: Captures precisely what is stated in text or represents meaning with minimal interpretation\\
        - SEMANTIC\_MODERATE: Requires some interpretation but maintains core meaning\\
        - SEMANTIC\_WEAK: Significant interpretation needed; meaning partially preserved\\
        - SEMANTIC\_NA: Not applicable\\
\\
    Abstraction Labels (select one):\\
        - ABSTRACTION\_BROADER: Represents a more general concept that includes text concepts\\
        - ABSTRACTION\_ALIGNED: Represents approximately the same scope and specificity as the text\\
        - ABSTRACTION\_NARROWER: Represents a more specific subset of text concepts\\
        - ABSTRACTION\_NA: Not applicable or the concepts were so different as to make abstraction comparison impossible\\
\\
\#\# Node Level Recall\\
For each node in the ground-truth graph, evaluate against the LLM's response:\\
\\
Response Evaluation\\
- Explicitly identify and quote ALL potentially corresponding nodes from LLM's response\\
- Apply these labels where applicable:\\
    Importance Labels (select one):\\
        - IMPORTANCE\_CORE: Ground-truth node represents a fundamental concept central to the causal structure\\
        - IMPORTANCE\_INTERMEDIATE: Ground-truth node serves as a key connection between central concepts\\
        - IMPORTANCE\_PERIPHERAL: Ground-truth node provides supplementary or contextual information\\
\\
    Presence Labels (select one):\\
        - PRESENCE\_STRONG\_MATCH: Core concept appears in response with only minor, inconsequential differences\\
        - PRESENCE\_WEAK\_MATCH: Core concept shares significant meaning with a response node but has notable differences\\
        - PRESENCE\_NO\_MATCH: No response node captures a similar core concept\\
\\
    Semantic Labels (select one):\\
        - SEMANTIC\_COMPLETE: Ground-truth concept fully captured with high fidelity, whether in single or multiple nodes\\
        - SEMANTIC\_PARTIAL: Core aspects captured but with some meaning loss or missing implications\\
        - SEMANTIC\_MINIMAL: Only basic or surface-level aspects of the concept captured\\
        - SEMANTIC\_NA: Not applicable\\
\\
    Abstraction Labels (select one):\\
        - ABSTRACTION\_BROADER: Represents a more general concept that includes the ground-truth node\\
        - ABSTRACTION\_ALIGNED: Represents approximately the same scope and specificity of the ground-truth node\\
        - ABSTRACTION\_NARROWER: Represents a more specific subset of the ground-truth node\\
        - ABSTRACTION\_NA: Not applicable or the concepts were so different as to make abstraction comparison impossible\\
\\
\# Edge Level Evaluation\\
\\
\#\# Edge Precision\\
For each edge (causal relationship) in the LLM's response, evaluate against both ground truth sources:\\
\\
1. Ground-Truth Graph Evaluation\\
- Explicitly identify and quote ALL potentially corresponding edges from ground-truth graph\\
- Apply these labels where applicable:\\
    Presence Labels (select one):\\
        - PRESENCE\_STRONG\_MATCH: Edge connects highly similar concepts as in ground-truth\\
        - PRESENCE\_WEAK\_MATCH: Edge connects somewhat similar concepts as in ground-truth\\
        - PRESENCE\_NO\_MATCH: No corresponding edge exists in ground-truth\\
\\
    Directionality Labels:\\
        - DIRECTION\_CORRECT: Direction of causality matches ground-truth\\
        - DIRECTION\_REVERSED: Direction of causality is opposite of ground-truth\\
        - DIRECTION\_NA: Not applicable or the concepts were so different as to make direction comparison impossible\\
\\
    Abstraction Labels:\\
        - ABSTRACTION\_ALIGNED: Edge represents similar scope of relationship as ground-truth\\
        - ABSTRACTION\_BROADER: Edge is substantially more general than ground-truth\\
        - ABSTRACTION\_NARROWER: Edge is substantially more specific than ground-truth\\
        - ABSTRACTION\_NA: Not applicable or the concepts were so different as to make abstraction comparison impossible\\

2. Ground-Truth Text Evaluation\\
- Explicitly quote ALL relevant supporting text that describes causal relationships\\
- Apply these labels where applicable:\\
    Evidence Labels (select one):\\
        - PRESENCE\_GRAPH\_ONLY: Causal relationship present in ground-truth graph (always select this if present)\\
        - PRESENCE\_EXPLICIT: Causal relationship directly stated in text (only if not in graph)\\
        - PRESENCE\_IMPLIED: Causal relationship can be reasonably inferred from text (only if not in graph)\\
        - PRESENCE\_NO\_MATCH: No text supports this causal relationship (only if not in graph)\\
\\
    Inference Labels (select one):\\
        - INFERENCE\_DIRECT: Relationship matches text's explicit causal claims\\
        - INFERENCE\_DERIVED: Relationship logically follows from text\\
        - INFERENCE\_STRETCHED: Relationship possible but weakly supported\\
        - INFERENCE\_NA: Not applicable or relationship does not exist\\
\\
    Abstraction Labels (select one):\\
        - ABSTRACTION\_ALIGNED: Matches the granularity of text's causal claims\\
        - ABSTRACTION\_BROADER: Generalizes multiple textual relationships\\
        - ABSTRACTION\_NARROWER: Specifies a subset of text's causal claims\\
        - ABSTRACTION\_NA: Not applicable or the concepts were so different as to make abstraction comparison impossible\\
\\
\#\# Edge Level Recall\\
For each causal relationship (edge) in the ground-truth graph, evaluate against the LLM's response:\\
\\
Response Evaluation\\
- Explicitly identify and quote ALL potentially corresponding causal relationships from LLM's response\\
- Apply these labels where applicable:\\
    Importance Labels (select one):\\
        Importance is based on how important it is to the ground-truth graph, regardless of whether it is present or accurately represented in the LLM's response.\\
\\
        - IMPORTANCE\_CENTRAL: A key causal relationship that drives main effects\\
        - IMPORTANCE\_CONNECTING: Links major causal chains together\\
        - IMPORTANCE\_AUXILIARY: Provides supplementary causal context\\

    Presence Labels (select one):\\
        - PRESENCE\_STRONG\_MATCH: Core concept appears in response with only minor, inconsequential differences\\
        - PRESENCE\_WEAK\_MATCH: Core concept shares significant meaning with a response node but has notable differences\\
        - PRESENCE\_NO\_MATCH: No response node captures a similar core concept\\
\\
    Directionality Labels (select one):\\
        - DIRECTION\_CORRECT: Causal relationship captured with correct direction\\
        - DIRECTION\_REVERSED: Causal relationship present but direction is reversed\\
        - DIRECTION\_UNCLEAR: Relationship present but direction is ambiguous\\
        - DIRECTION\_MISSING: Relationship entirely absent from response\\
\\
    Abstraction Labels (select one):\\
        - ABSTRACTION\_ALIGNED: One-to-one relationship match at similar level of detail\\
        - ABSTRACTION\_BROADER: Edge is substantially more general than ground-truth\\
        - ABSTRACTION\_NARROWER: Edge is substantially more specific than ground-truth\\
        - ABSTRACTION\_NA: Not applicable or the concepts were so different as to make abstraction comparison impossible\\
\\
\# Expected Output Format\\
The output should be in YAML format. Only include the evaluation sections that are being evaluated - omit other sections entirely. For example, if only evaluating node precision, only the node\_precision\_evaluations section should be present. However, within the required evaluation sections, be sure to always include the Not Applicable labels rather than omitting them.\\

\begin{verbatim}
```yaml
\end{verbatim}
\# If evaluating node precision:\\
node\_precision\_evaluations:\\
  - node\_number: \verb|<integer>|\\
    graph\_evaluation:\\
      presence\_label: \verb|<PRESENCE_LABEL>|\\
      semantic\_label: \verb|<SEMANTIC_LABEL>|\\
      abstraction\_label: \verb|<ABSTRACTION_LABEL>|\\
    text\_evaluation:\\
      presence\_label: \verb|<PRESENCE_LABEL>|\\
      semantic\_label: \verb|<SEMANTIC_LABEL>|\\
      abstraction\_label: \verb|<ABSTRACTION_LABEL>|\\
\\
\# If evaluating node recall:\\
node\_recall\_evaluations:\\
  - node\_number: \verb|<integer>|\\
    importance\_label: \verb|<IMPORTANCE_LABEL>|\\
    presence\_label: \verb|<PRESENCE_LABEL>|\\
    semantic\_label: \verb|<SEMANTIC_LABEL>|\\
    abstraction\_label: \verb|<ABSTRACTION_LABEL>|\\
\\
\# If evaluating edge precision:\\
edge\_precision\_evaluations:\\
  - edge\_number: \verb|<integer>|\\
    graph\_evaluation:\\
      presence\_label: \verb|<PRESENCE_LABEL>|\\
      directionality\_label: \verb|<DIRECTION_LABEL>|\\
      abstraction\_label: \verb|<ABSTRACTION_LABEL>|\\
    text\_evaluation:\\
      presence\_label: \verb|<PRESENCE_LABEL>|\\
      inference\_label: \verb|<INFERENCE_LABEL>|\\
      abstraction\_label: \verb|<ABSTRACTION_LABEL>|\\
\\
\# If evaluating edge recall:\\
edge\_recall\_evaluations:\\
  - edge\_number: \verb|<integer>|\\
    importance\_label: \verb|<IMPORTANCE_LABEL>|\\
    presence\_label: \verb|<PRESENCE_LABEL>|\\
    directionality\_label: \verb|<DIRECTION_LABEL>|\\\\
    abstraction\_label: \verb|<ABSTRACTION_LABEL>|\\
\verb|```|
\end{tcolorbox}

\end{document}